\documentclass[runningheads]{llncs}

\usepackage{eccv}

\usepackage{eccvabbrv}

\usepackage{graphicx}
\usepackage{colortbl}
\usepackage{booktabs}
\usepackage{algorithm}
\usepackage{algpseudocode}
\usepackage{amsmath,amssymb}
\usepackage{pifont}
\usepackage{tcolorbox}
\usepackage{microtype}
\usepackage{enumitem}
\usepackage{subcaption}
\usepackage{wrapfig}
\usepackage{xcolor}
\usepackage{makecell}

\usepackage[accsupp]{axessibility}  

\usepackage{hyperref}
\usepackage{multirow}
\usepackage{orcidlink}

\newcommand{\ours}{\textsc{VisionCoach}\xspace}

\newcommand{\selector}{\textsc{VP-Selector}\xspace}

\newcommand{\reasoner}{\textsc{ST-Reasoner}\xspace}

\begin{document}




\title{\ours{}: Reinforcing Grounded Video Reasoning via Visual-Perception Prompting} 

\titlerunning{\ours{}}




\author{Daeun Lee\orcidlink{0009-0006-2717-6964} \and
Shoubin Yu\orcidlink{0009-0006-1670-0054} \and
Yue Zhang\orcidlink{0000-0003-2153-6536} \and
Mohit Bansal\orcidlink{0009-0009-2965-5354}}

\authorrunning{D.~Lee et al.}

\institute{University of North Carolina, Chapel Hill, USA\\
\email{\{daeun,shoubin,yuezhan,mbansal\}@cs.unc.edu}\\
\url{https://visioncoach.github.io/}}

\maketitle
\begin{abstract}
Video reasoning requires models to locate and track question-relevant evidence across frames. 
While reinforcement learning (RL) with verifiable rewards improves accuracy, it still struggles to achieve reliable spatio-temporal grounding during the reasoning process. 
Moreover, improving grounding typically relies on scaled training data or inference-time perception tools, which increases annotation cost or computational cost. 
To address this challenge, we propose \ours{}, an input-adaptive RL framework that improves spatio-temporal grounding through visual prompting as training-time guidance. 
During RL training, visual prompts are selectively applied to challenging inputs to amplify question-relevant evidence and suppress distractors. The model then internalizes these improvements through self-distillation, enabling grounded reasoning directly on raw videos without visual prompting at inference.
\ours{} consists of two components: (1) Visual Prompt Selector, which predicts appropriate prompt types conditioned on the video and question, and (2) Spatio-Temporal Reasoner, optimized with RL under visual prompt guidance and object-aware grounding rewards that enforce object identity consistency and multi-region bounding-box IoU.
Extensive experiments demonstrate that \ours{} achieves state-of-the-art performance across diverse video reasoning, video understanding, and temporal grounding benchmarks (V-STAR,  VideoMME, World-Sense, VideoMMMU, PerceptionTest, and Charades-STA), while maintaining a single efficient inference pathway without external tools. 
Our results highlight the effectiveness of training-time visual prompting as a lightweight mechanism for improving grounded video reasoning.

\end{abstract}
\section{Introduction}

\begin{figure}[t]
    \centering
    \includegraphics[width=0.95\linewidth]{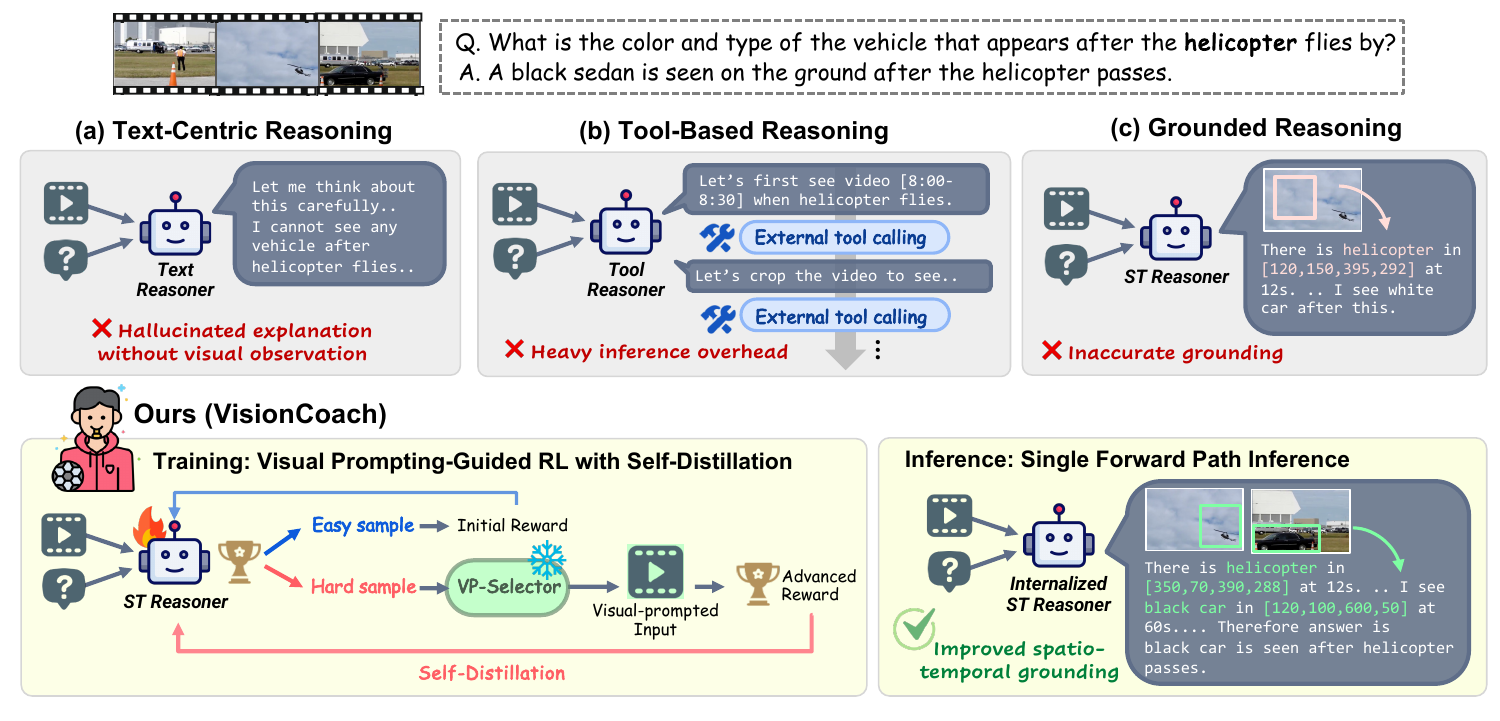}
    \caption{
    \textbf{Comparison with previous video reasoning approaches and \ours{}.}
    Comparing with other video reasoning models, \ours{} leverages visual-prompt-guided RL with self-distillation to internalize improved spatio-temporal grounding behaviors induced by visual guidance.
    At inference time, it maintains a single forward-pass reasoning while achieving enhanced grounding performance. 
    }
    \label{fig:teaser}
\end{figure}

Recent advances in reinforcement learning (RL) with verifiable rewards~\cite{liu2024deepseek} have begun to improve multimodal reasoning and are increasingly applied to video reasoning~\cite{cheng2025video,videomme,videommmu,deng2025scivideobench}.
Despite recent progress, existing video reasoning approaches still struggle to achieve reliable spatio-temporal grounding throughout the reasoning process. 
Text-centric video reasoning models~\cite{videor1,wang2025videorft,lee2025video} (\cref{fig:teaser}~(a)) often generate hallucinated explanations driven by language priors rather than faithful visual observations. 
Visual tool-calling approaches~\cite{videochatr15,videocom,ge2025framemind,thinkingwvideo,meng2025open,zeng2026video} (\cref{fig:teaser}~(b)) improve grounding by invoking external perception tools such as temporal clipping or zoom-in. 
While these tools can retrieve relevant evidence, the reasoning model itself does not internalize grounded perception and remains dependent on external modules at inference time. 
Recent grounded reasoning models~\cite{meng2025open} (\cref{fig:teaser}~(c)) attempt to integrate spatio-temporal evidence within a single model by interleaving grounding and reasoning. 
Nevertheless, grounding remains unreliable, frequently producing inaccurate object references or hallucinated bounding boxes that propagate errors during reasoning.

At the core of this issue is the absence of mechanisms that enforce alignment between intermediate reasoning steps and spatio-temporal evidence.
In practice, improving grounding typically requires either scaling training data or additional perception modules (e.g., cropping image regions or trimming the video) at inference time.
However, dense annotations across diverse video datasets are expensive to obtain, while inference-time tools increase computational overhead.
These solutions improve grounding only through heavier external intervention, rather than enhancing the model’s intrinsic perception behavior.
Instead, we shift the focus from data scaling and inference-time intervention to \textbf{training-time guidance}, aiming to internalize grounded reasoning behavior while preserving a lightweight inference pipeline.

To this end, we propose \ours{}, an input-adaptive RL framework that uses visual prompts as a training-time vision coach to improve spatio-temporal grounding while enabling inference directly on raw videos. 
The key idea is to selectively apply visual prompts to challenging inputs during RL training to strengthen grounding, and to internalize these improvements through self-distillation so that visual prompting is no longer required at inference. 
Instead of relying solely on implicit feature-space attention, \ours{} performs reasoning-driven perception control at the input level, amplifying question-relevant evidence and suppressing distractors during training.

Specifically, \ours{} consists of two components: 
(i) Visual Prompt Selector (\selector), which predicts an appropriate visual prompt type conditioned on the video and question to provide adaptive perception guidance for challenging inputs, and 
(ii) Spatio-Temporal Reasoner (\reasoner), which is optimized with RL under visual prompt-guided perception and grounding-aware rewards.
Before training \reasoner, we first construct a visual prompting candidate dataset using a proxy reasoner and train \selector with SFT. 
During \reasoner training, hard samples are identified based on initial rewards, and the frozen \selector predicts visual prompt types that guide the \reasoner to attend to relevant regions and moments, producing more grounded reasoning trajectories.
To further strengthen grounded reasoning, we introduce an object-aware spatial grounding reward that enforces object identity consistency and average IoU across multiple predicted bounding boxes, encouraging accurate multi-object grounding and temporally consistent reasoning.
To remove prompt dependency at inference, we employ self-distillation, where the prompted input serves as a coach to guide the policy model, enabling grounded perception behavior to be internalized. 
At inference, the model performs reasoning directly on raw videos with a single forward pass, without visual prompting.
Together, these designs provide localized perception guidance during training while keeping a simple and efficient inference pipeline.

We evaluate \ours{} on the (1) spatio-temporal reasoning benchmark V-STAR~\cite{vstar}, (2) general video understanding benchmarks (VideoMME~\cite{videomme}, WorldSense~\cite{hong2025worldsense}, VideoMMMU~\cite{videommmu}, and PerceptionTest~\cite{perceptiontest}), and (3) temporal grounding benchmark (Charades-STA~\cite{gao2017tall}). 
On V-STAR, \ours{} surpasses GPT-4o and improves over Qwen2.5-VL-7B by +15.0\% mAM and +25.1\% mLGM, establishing new state-of-the-art performance. 
In general video understanding and temporal grounding benchmarks, it consistently outperforms prior open-source VideoLLMs, demonstrating strong performance in long-video reasoning and perception-oriented understanding tasks. 
We further provide comprehensive analyses, including spatio-temporal attention map visualization, component-wise ablation, generalization effect of \selector across different backbone models, and statistics of adaptive visual prompting. 

Our contributions are summarized as follows: 
\begin{itemize}[leftmargin=*, itemsep=1pt, topsep=1pt]
\item We propose an input-adaptive RL framework for video reasoning that explicitly guides spatio-temporal grounding through training-time visual prompting and self-distillation, enabling the reasoning model to internalize grounded perception without requiring visual prompts at inference time.
\item We design an object-aware spatial grounding reward that incorporates object identity consistency and multi-region bounding-box IoU to better support spatial-grounded reasoning.
\item We introduce a visual prompt selector with a proxy-reasoner-based data construction pipeline to predict appropriate visual prompts for video QA inputs. 
\item Extensive experiments and analyses demonstrate that \ours{} achieves state-of-the-art performance across a wide range of video reasoning, video understanding, and temporal grounding benchmarks.
\end{itemize}

\section{Related Work}

\noindent\textbf{Video Reasoning.}
Video understanding has advanced rapidly with large multimodal models~\cite{li2024llava,bai2025qwen3,li2023blip}, enabling complex video QA and reasoning across diverse areas~\cite{videomme,videommmu,perceptiontest,vstar,hong2025worldsense,wu2024star, deng2025motion, yu2026and}.
Despite progress, reliable \emph{grounded} video reasoning remains challenging because models must track object states, events, and interactions over time.
A growing line of work applies reinforcement learning with verifiable or rule-based rewards to strengthen multimodal reasoning~\cite{deepseek_r1,videochatr1,videochatr15,thinkingwvideo,meng2025open,zeng2026video,wang2025video,chen2025scaling, cheng2026graphthinker}, but existing approaches often either (i) remain text-centric and hallucinate evidence, or (ii) rely on iterative tool operations at inference time (e.g., progressive ROI localization), introducing overhead and leaving limited explicit control over dense spatio-temporal evidence.
Our work targets this gap by using \emph{training-time} perception control to improve grounding. 

\noindent\textbf{Visual Prompting.}
Visual prompting~\cite{wu2024visual} augments inputs with lightweight visual cues (e.g., boxes, masks, points, scribbles, overlays) to steer attention and improve localized understanding without heavy architectural changes.
Early and concurrent studies show that simple edits in pixel space, such as drawing a red circle around an object, can reliably direct VLM attention~\cite{redcircle}.
Recent work expands visual prompting to general vision understanding in MLLMs~\cite{vipllava, vincontext,choudhury2024video,gu2025thinking}.
For video settings, prompting has also been used to improve temporal grounding by adding structured cues such as per-frame numbering~\cite{numpro,zhang2025vtimecot}.
Meanwhile, learned or automated prompting methods aim to \emph{select} or \emph{retrieve} effective prompts conditioned on the input, improving robustness and usability~\cite{autov,tvp}.
In \ours{}, we use adaptive prompting \emph{only during RL training} and then internalize the benefit so inference no longer depends on prompting.

\noindent\textbf{Model Distillation.}
Knowledge distillation transfers behavior from a stronger teacher to a student model, improving efficiency and robustness~\cite{hinton2015distill}.
Beyond classical teacher-student training, self-distillation and iterative teacher replacement can further improve generalization without additional labels~\cite{bornagain}.
Distillation has also been used in sequential decision making (policy distillation) to compress or unify behaviors learned via RL~\cite{policy_distill}.
In \ours{}, self-distillation helps the ST-Reasoner \emph{internalize} the grounding improvements induced by visual prompting guided training trajectories, enabling a single, prompt-free inference.

\begin{figure*}[t]
    \centering
    {
    \includegraphics[width=\textwidth,height=0.41\textheight]{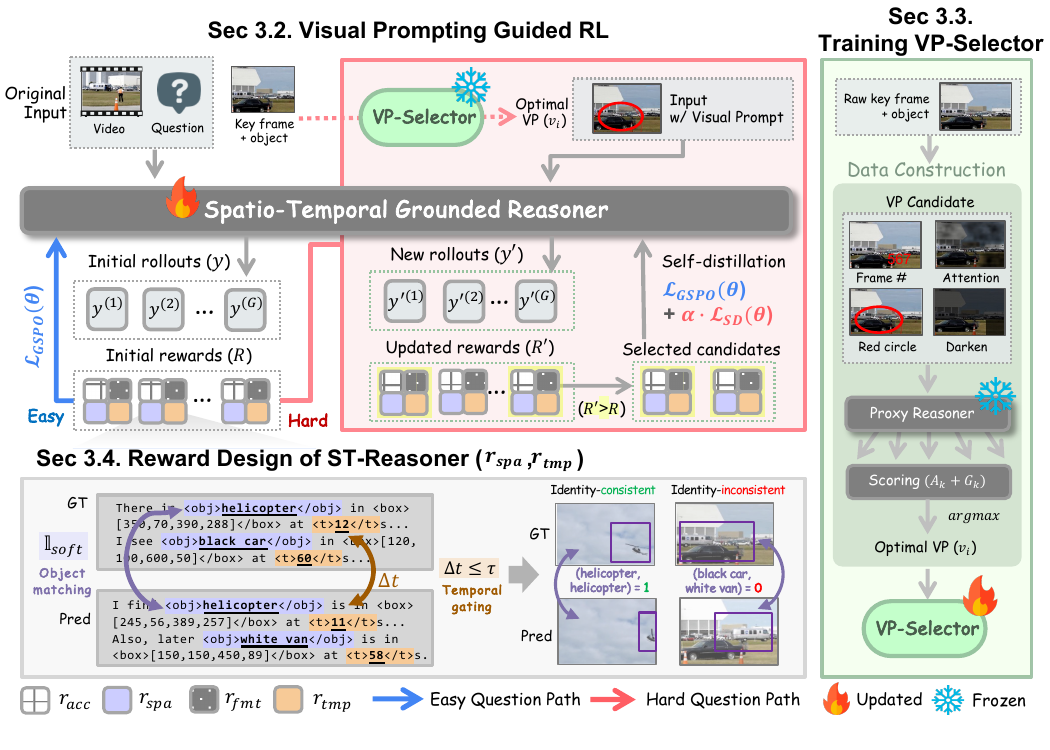}}
    \caption{
    \small \textbf{Detailed Architecture of \ours{}. } 
    We introduce \ours{}, a visual-prompt-guided RL framework for training a spatio-temporally grounded reasoner (\cref{sec:method-pipeline}). 
    The framework includes Visual Prompt Selector (\selector) that predicts optimal visual prompts for challenging inputs (\cref{sec:method-vpselector}), 
    and object-aware spatial grounding rewards that enforce object identity consistency and multiple predicted bounding boxes (\cref{sec:method-reward}). 
    }
    \label{fig:method}
\end{figure*}

\section{Method}
As shown in~\cref{fig:method}, we propose \ours{}, an input-adaptive RL framework that improves spatio-temporal grounding in video reasoning through visual prompting guidance.
We begin by formalizing spatio-temporal grounded reasoning and presenting the motivation for our approach (\cref{sec:method-problem}). 
Next, we introduce the overall training pipeline of \ours{} in \cref{sec:method-pipeline}, which integrates visual prompting with RL and self-distillation to encourage grounded reasoning. 
Finally, we describe the key components in detail: the Visual Prompt Selector (\cref{sec:method-vpselector}), which predicts input-adaptive visual prompts, followed by the reward design of Spatio-Temporal Reasoner  (\cref{sec:method-reward}), which learns grounded reasoning from prompt-guided training signals and grounding-aware rewards.

\subsection{Problem Statement and Motivation}\label{sec:method-problem}
Given an input video $x$ and a question $q$, the goal of video QA is to generate a grounded reasoning trajectory $y$ and predict the final answer $a$ over complex and dynamic visual content.
Unlike text-based reasoning (\cref{fig:teaser} (a)), our reasoning requires a policy model $\pi_\theta$ to integrate explicit spatio-temporal grounding, identifying when and where relevant evidence occurs while reasoning.

\begin{wrapfigure}{r}{0.5\columnwidth} 
    \begin{minipage}{\linewidth}
        \centering
        \includegraphics[width=\linewidth]{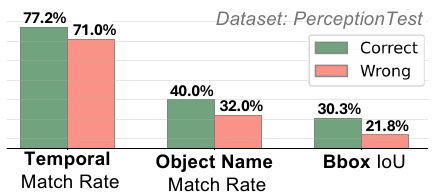}
        \captionof{figure}{\textbf{Grounding effect on answering.}}
        \label{fig:proof}
        \resizebox{\linewidth}{!}{%
        \begin{tabular}{lcccc}
            \toprule
            \textbf{Models} & \textbf{\begin{tabular}[c]{@{}c@{}}Raw key\\frame\end{tabular}} & \textbf{Darken} & \textbf{Red circle} & \textbf{\begin{tabular}[c]{@{}c@{}}Oracle\\(Any-corr)\end{tabular}} \\
            \midrule
            Qwen2.5-VL       & 52.2 & 43.3 & 51.5 & \textbf{70.8} \\
            Gemini-2.5-Flash & 47.4 & 50.4 & 49.3 & \textbf{75.2} \\
            \bottomrule
        \end{tabular}
        }
        \captionof{table}{\textbf{Effect of visual prompting.}} 
        \label{table:effect_vp}
    \end{minipage}
    \vspace{-.5cm}
\end{wrapfigure}

\noindent\textbf{Motivation.} 
To better understand the relationship between grounding and answering performance, we analyze model behaviors on the PerceptionTest subset. 
As shown in \cref{fig:proof}, correctly answered samples consistently exhibit higher temporal alignment, object identity matching, and spatial IoU compared to incorrectly answered ones, indicating that accurate spatio-temporal grounding  is correlated with correct answering.
We further investigate how different visual prompts affect answering performance. 
As shown in \cref{table:effect_vp}, different prompts lead to different results, suggesting that selecting an appropriate visual prompt is crucial for effective answering~\cite{numpro,autov,zhang2025vtimecot}. 
Notably, the oracle prompt selection achieves significantly higher accuracy, indicating that adaptive perceptual guidance can substantially improve downstream answering when the appropriate prompt is applied. 
Motivated by these, we propose an RL framework that enables input-adaptive visual prompting to achieve reliable grounding and answering.

\subsection{\ours{}: VP-Guided RL with Self-Distillation}\label{sec:method-pipeline}

We introduce \ours, an RL framework that integrates \selector{} and \reasoner{} to enable input-adaptive visual guidance during training. 
We describe each component in detail in \cref{sec:method-vpselector} (\selector) and \cref{sec:method-reward} (Reward design for \reasoner).
The \reasoner{} $\pi_\theta$ is optimized using GSPO~\cite{gspo}, starting from a cold-start initialization and trained on video–question pairs $(x,q)$. 
For each input, $G$ reasoning trajectories are sampled and evaluated with grounding-aware rewards.
Visual prompts are selectively applied to challenging inputs during RL to strengthen spatio-temporal grounding, and their benefits are internalized through self-distillation, eliminating the need for visual prompting at inference. 
The overall training procedure is shown in \cref{alg:qad-gspo-vp-idl}.

\noindent\textbf{Input-adaptive hard sample identification.}
Given an input $(x_i, q_i)$, we first perform $G$ initial rollouts using the current policy.
Let $\{R_i^{(g)}\}_{g=1}^{G}$ denote the resulting overall rewards.
We compute the average reward $\bar{R}_i$ and classify the sample as hard if $\bar{R}_i < k$, where $k$ is a predefined threshold for hard sample filtering.
This input-adaptive mechanism determines whether additional visual guidance is necessary for each example, allowing more targeted visual prompting to help with grounding.

\begin{algorithm}[t]
\centering
\caption{\small \ours{}: Visual Prompt Guided RL with Self-Distillation}
\label{alg:qad-gspo-vp-idl}
\begin{algorithmic}[1]

\Require Dataset $\mathcal{D}=\{(x_i,q_i)\}_{i=1}^N$, \reasoner $\pi_\theta$, trained \selector, reward $r$, number of rollouts $G$, hard question threshold $k$, self-distillation weight $\alpha$.

\For{each $(x_i,q_i)\in\mathcal{D}$} 
  \For{$g \gets 1$ to $G$}
    \State $y_i^{(g)} \sim \pi_\theta(\cdot \mid x_i,q_i),\;
    R_i^{(g)} \gets r(x_i,y_i^{(g)})$
    \hfill \textcolor{gray}{\footnotesize\texttt{//* Conduct initial rollout}}
  \EndFor
  \State $\bar{R}_i \gets \frac{1}{G}\sum_{g=1}^G R_i^{(g)}$,\; 
    $\mathbb{I}_i \gets \mathbf{1}[\bar{R}_i < k]$ 
    
  \If{$\mathbb{I}_i = 1$}
      \hfill \textcolor{gray}{\footnotesize\texttt{//* Generate VP guidance}}
    \State $v_i \gets \selector(x_i,q_i)$ 
    \State $(x_i',q_i') \gets (\texttt{VP}(x_i;v_i),\; q_i \oplus \texttt{hint}(v_i))$ 

  \For{$g \gets 1$ to $G$}
    \State $y_i'^{(g)} \sim \pi_\theta(\cdot \mid x_i',q_i')$, \;
     $R_i'^{(g)} \gets r(x_i',y_i'^{(g)})$ 
  \EndFor

  \State $\bar{R}_{i}' \gets \frac{1}{G}\sum_{g=1}^G R_{i}'^{(g)}$, \; 
  $C_i \gets \{g : R_{i}'^{(g)} > \bar{R}_{i}\}$

  \If{$C_i \neq \emptyset$}       \hfill \textcolor{gray}{\footnotesize\texttt{//* Conduct self-distillation}}
    \State $S_i \gets \texttt{TopN}(\{R_{i,\mathrm{ans}}'^{(g)}\}_{g\in C_i})$ 

\State $\mathcal{L}_{\mathrm{SD}}(\theta) \gets
-\frac{1}{|S_i|}
\sum_{j\in S_i}
\frac{1}{|y_{i,j}'|}
\sum_{t=1}^{|y_{i,j}'|}
\log \pi_\theta(y'_{i,j,t}\mid y'_{i,j,<t},x_i',q_i')$
  \State $\mathcal{L}(\theta) \gets \mathcal{L}_{\mathrm{GSPO}}(\theta) + \alpha\cdot \mathbb{I}_i \cdot \mathcal{L}_{\mathrm{SD}}(\theta)$
\EndIf
\Else
\State $\mathcal{L}(\theta) \gets \mathcal{L}_{\mathrm{GSPO}}(\theta)$ 
\EndIf
\EndFor
\end{algorithmic}
\end{algorithm}

\noindent\textbf{Visual prompting guidance generation.}
For hard samples, we feed $(x_i, q_i)$ into the trained \selector, which is elaborated in~\cref{sec:method-vpselector} to obtain the optimal visual prompt ~(such as darken), which is denoted as $v_i$.
We then apply this prompt to the key frames of $x_i$ to construct a visual-prompted input $x_i'$, and append a textual \texttt{hint} describing the applied visual prompting to the original question $q_i$ and make $q_i'$.
The visual prompt introduces localized cues that facilitate spatio-temporal grounding.
For example, suppressing irrelevant regions can emphasize key object areas and improve spatial reasoning.
Using the prompted input $(x_i', q_i')$, we perform another $G$ rollouts and obtain new reasoning $y_i'$ and updated overall rewards $\{R_i'^{(g)}\}_{g=1}^{G}$.
We expect improved reasoning trajectories and higher rewards due to the enhanced grounding signals provided by visual prompting.
Detailed ablation studies on the choice of reward for candidate selection are provided in the Appendix.

\noindent\textbf{Self-distillation.}
When visual prompting yields improved rewards, we further reinforce this improved reasoning through self-distillation.
Specifically, after performing rollouts on the visual-prompted input, we identify the subset of rollouts whose overall rewards exceed the average reward of the original rollouts, i.e., 
$C_i = \{ g \mid R_i'^{(g)} > \bar{R}_i \}$.
Among these candidates, we select the top N rollouts $S_i$ based on answering rewards. 
If no such candidates exist (i.e., $C_i=\emptyset$), we skip the self-distillation step for the current sample.
We then apply token-level negative log-likelihood (NLL) to the selected reasoning: 
\begin{equation}
\mathcal{L}_{\mathrm{SD}}(\theta)
=
-\frac{1}{|S_i|}
\sum_{j\in S_i}
\frac{1}{|y_{i,j}'|}
\sum_{t=1}^{|y_{i,j}'|}
\log \pi_\theta(y'_{i,j,t}\mid y'_{i,j,<t},x_i',q_i').
\end{equation}

This objective encourages the policy to internalize high-reward reasoning trajectories generated under visual guidance.
Accordingly, the final training objective is defined as:
\begin{equation}
\mathcal{L}_{\mathrm{GSPO}}(\theta)
+
\alpha \, \mathbb{I}_i \, \mathcal{L}_{\mathrm{SD}}(\theta),
\end{equation}
where $\mathbb{I}_i$ is an indicator function that equals 1 if sample $i$ is identified as a hard sample, and 0 otherwise.
By repeatedly reinforcing improved trajectories, the model progressively internalizes the grounding behaviors induced by visual prompting, enabling self-evolving and more robust spatio-temporal reasoning without requiring visual prompts at inference time.

\subsection{\selector: Learning to Provide Visual Guidance}\label{sec:method-vpselector}
We now introduce \selector, which is used to train the policy model \reasoner within our input-adaptive RL framework. 
Our \selector is designed to predict an input-adaptive visual prompt conditioned on the video–question pair, enabling targeted visual guidance during RL training for hard examples.
Given an input $(x, q)$, the \selector selects an appropriate visual prompt $v_i$ from a candidate pool, where each prompt provides a different form of perceptual guidance.
As shown in \cref{fig:method} right, we first construct a training dataset using proxy reasoners~\cite{gemini2,gpt4o,bai2025qwen3}  
to estimate the effectiveness of candidate prompts and train the small VLM to predict the most suitable visual prompt.

\noindent\textbf{Training data collection.}
Since defining a gold visual prompt across models is challenging, we collect $(x, v_i)$ pairs using multiple proxy reasoners to capture general prompt effectiveness patterns.  
We first define a visual prompt candidate $\mathcal{V}=\{v_1, v_2, \dots\}$ by applying diverse visual prompting to key frames and objects from $x$. Specifically, we consider red circles~\cite{redcircle}, attention-based prompts~\cite{api}, frame numbering~\cite{numpro}, and darkening as potential visual guidance. 
For each candidate prompt $v_k$, we generate a viusal-prompted input $x_k'=\texttt{VP}(x;v_k)$ and obtain reasoning outputs $y_k=\mathcal{R}(x_k',q)$ using multiple proxy reasoners $\mathcal{R}$ ~\cite{comanici2025gemini,gpt4o,bai2025qwen3}. 
We compute binary answer accuracy and grounding scores $(A_k,G_k)$, average them across proxy reasoners, and select the optimal prompt as

\begin{equation}
    v_i=\arg\max_{v_k\in\mathcal{V}} (A_k + G_k).
\end{equation}

The resulting $(x,q,v_i)$ pairs are used to train the VP-Selector to predict the optimal prompt conditioned on the input video and question. 
Please refer to the Appendix for more details about the collected data.

\noindent\textbf{Training.}
Given the collected dataset $\mathcal{D}=\{(x,q,v_i)\}$, we train the \selector as a lightweight VLM classifier~\cite{bai2025qwen2} with LoRA~\cite{hu2022lora}. We cast prompt selection as a $|\mathcal{V}|$-way prediction problem and optimize the selector using a supervised objective. Concretely, we format the input $(x,q)$ as an instruction to choose one method from $\mathcal{V}$ and train the model to generate the corresponding prompt label as a single-token/short-string response using token-level cross-entropy loss. The \selector is frozen when incorporated in the RL framework.
Additional details are provided in the Appendix.

\subsection{Reward Design of \reasoner}\label{sec:method-reward}
We design reward functions to train \reasoner with GSPO. 
Specifically, we employ four reward components: 
(1) answer accuracy, 
(2) format correctness,  
(3) temporal grounding, and
(4) object-aware spatial grounding.
Among them, the object-aware spatial grounding reward is newly introduced in this work, while the remaining rewards follow prior work~\cite{meng2025open}. The overall reward is defined as 

\begin{equation}
r(x,y)=r_{\text{acc}}(x,y)+r_{\text{fmt}}(x,y)+r_{\text{tmp}}(x,y)+r_{\text{spa}}(x,y),
\end{equation}
and the rewards are group-normalized across rollouts to compute advantages used for GSPO updates.

\noindent\textbf{Accuracy reward ($r_{\text{acc}}$).}
Following~\cite{meng2025open}, we define task-specific accuracy rewards depending on the supervision type. 
For multiple-choice questions, the reward is binary correctness. For open-ended questions, we compute textual similarity between the predicted and ground-truth answers using ROUGE. 
For spatial grounding tasks, the reward is given by the visual IoU between predicted and GT bounding boxes, while for temporal grounding tasks we use temporal IoU between predicted and GT time intervals. 
This design provides a unified accuracy signal that adapts to heterogeneous task formats.

\noindent\textbf{Format reward ($r_{\text{fmt}}$).}
We assign a binary reward that evaluates whether the generated output follows the required structured format, including the correct use of \texttt{<think>} and \texttt{<answer>} tags as well as valid \texttt{<obj>}, \texttt{<box>}, and \texttt{<t>} annotations. Outputs that satisfy the format receive a reward of $1$, and $0$ otherwise.

\noindent\textbf{Temporal grounding reward ($r_{\text{tmp}}$).}
Following prior work~\cite{meng2025open}, we set the temporal grounding reward to force the model locate correct temporal boundaries for visual evidence.
Let $\mathcal{T}^{\text{gt}} \subset \mathbb{R}$ denote the set of annotated ground-truth temporal positions.
Depending on the training dataset, annotations are provided either as discrete timestamps or as temporal intervals.
When interval annotations are available, we denote the ground-truth interval set as
$\mathcal{S}^{\text{gt}} = \{[s_\ell^{\text{gt}}, e_\ell^{\text{gt}}]\}_{\ell=1}^{L}$.
From the generated reasoning $y$, we parse $M$ predicted timestamps $\{t_m\}_{m=1}^{M}$.
For each $t_m$, we first match it to the closest ground-truth temporal position:
\[
\tilde{t}_m = \arg\min_{t \in \mathcal{T}^{\text{gt}}} |t_m - t|,
\qquad
\Delta t_m = |t_m - \tilde{t}_m|.
\]

Then, we define the temporal reward as \begin{equation} r_{\text{tmp}}(x,y)= \frac{1}{M}\sum_{m=1}^{M} r_m, \end{equation} where 
\begin{equation} 
r_m =
\begin{cases}
1, & \text{if } \exists \ell:\; t_m \in [s_\ell^{\text{gt}}, e_\ell^{\text{gt}}], \\
\exp\!\Big(-\frac{\Delta t_m^2}{2\sigma^2}\Big), & \text{otherwise}.
\end{cases}
\end{equation}

\noindent\textbf{Object-aware spatial grounding reward ($r_{\text{spa}}$).}
Here, we further design an object-aware spatial grounding reward to better support grounded reasoning. 
Prior work~\cite{meng2025open} computes spatial rewards by selecting only the predicted bounding box with the maximum IoU, without considering object identity.
However, we observe that this design encourages single-box predictions and object-agnostic hallucinations, as it does not enforce consistency between the predicted objects and the grounded region.
As shown in the bottom of~\cref{fig:method}, to promote accurate object identification and multi-object grounding, we introduce a spatial reward that jointly considers soft object identity matching and all predicted box IoUs under temporal gating.

From the generated reasoning $y$, we parse predicted objects and bounding boxes to construct spatio-temporal tuples
$\{(o_m, t_m, b_m)\}_{m=1}^{M}$, 
where $o_m$ denotes the predicted object name and $b_m$ denotes the predicted bounding box.
We denote the ground-truth object set and their bounding boxes at the matched keyframe $\tilde{t}_m$ as
$\mathcal{O}^{\text{gt}}(\tilde{t}_m)$ and
$\mathcal{B}^{\text{gt}}(\tilde{t}_m)$, respectively.
Using soft identity matching $\mathbb{I}_{\text{soft}}(o_m,o)$, we define the identity-consistent object set as
\begin{equation}
\mathcal{O}_m^{\text{match}}
=
\left\{
o \in \mathcal{O}^{\text{gt}}(\tilde{t}_m)
\;\middle|\;
\mathbb{I}_{\text{soft}}(o_m,o)=1
\right\}.
\end{equation}

Here $\mathbb{I}_{\text{soft}}(o_m,o)=1$ if the predicted object name $o_m$ and the ground-truth name $o$ satisfy one of the following conditions:
(i) exact string match,
(ii) substring inclusion (i.e., $o_m \subset o$ or $o \subset o_m$);
otherwise $\mathbb{I}_{\text{soft}}(o_m,o)=0$.

Moreover, since spatial grounding is meaningful only when temporal localization is sufficiently accurate, we apply \textbf{temporal gating} based on the deviation $\Delta t_m$ defined above.
We retain only predictions that satisfy temporal alignment and identity consistency:

\[
\mathcal{I}
=
\left\{
m \;\middle|\;
\Delta t_m \le \tau
\;\wedge\;
|\mathcal{O}_m^{\text{match}}|>0
\right\},
\]
where $\tau$ is a predefined temporal tolerance threshold.
The spatial reward is then computed by averaging IoU scores over all valid predictions:

\begin{equation}
r_{\text{spa}}(x,y)
=
\frac{1}{|\mathcal{I}|}
\sum_{m\in\mathcal{I}}
\frac{1}{|\mathcal{O}_m^{\text{match}}|}
\sum_{o\in\mathcal{O}_m^{\text{match}}}
\max_{b^{\text{gt}}\in\mathcal{B}^{\text{gt}}(\tilde{t}_m)}
\mathrm{IoU}(b_m,b^{\text{gt}}).
\end{equation}

\section{Experiments}
\begin{table*}[t]
  \centering
  \caption{\small \textbf{Performance on the \textbf{V-STAR} benchmark.}
  \ours{} achieves strong spatio-temporal reasoning across overall dimensions.}
  \renewcommand{\arraystretch}{0.85}
  \resizebox{0.95\textwidth}{!}{
  \scriptsize
  \begin{tabular}{lccccccc}
    \toprule[0.15em]
    \textbf{Model} & 
    \textbf{What} &
    \multicolumn{2}{c}{\textbf{When (Temporal IoU)}} &
    \multicolumn{2}{c}{\textbf{Where (Spatial IoU)}} & 
    \multicolumn{2}{c}{\textbf{Overall}} \\
    \cmidrule(lr){2-2} \cmidrule(lr){3-4} \cmidrule(lr){5-6} \cmidrule(lr){7-8}
                  & 
     \textbf{Acc} & 
     \textbf{Chain1} & 
     \textbf{Chain2} &
     \textbf{Chain1} &
     \textbf{Chain2} & 
     \textbf{\textbf{mAM}} &
     \textbf{\textbf{mLGM}} \\
    \midrule
    \rowcolor{red!8} \multicolumn{8}{l}{\emph{Proprietary Models}} \\
    Gemini-2-Flash~\cite{gemini2}  & 53.0 & 24.5 & 23.8 & 4.6 & 2.2 & 26.9 & 35.6 \\
    GPT-4o~\cite{gpt4o} & 60.8 & 16.7 & 12.8 & 6.5 & 3.0 & 26.8 & 38.2 \\
    \midrule
    \rowcolor{blue!8} \multicolumn{8}{l}{\emph{Tool-calling Methods}} \\
    VideoZoomer~\cite{ding2025videozoomer} & 47.7 & 11.1 & 3.7 & 2.0 & 0.5 & 18.8 & 24.6 \\
    Conan~\cite{ouyang2026conan} & 54.5 & 14.4 & 14.9 & 4.4 & 0.6 & 23.9 & 32.4 \\
    Video-o3~\cite{videoo3} & 29.7 & 5.5 & 23.1 & 0.5 & 0.0 & 13.3 & 15.4 \\
    \midrule
    \rowcolor{green!8} \multicolumn{8}{l}{\emph{Tool-free Methods}} \\
    TRACE~\cite{guo2024trace} & 17.6 & 19.1 & 17.1 & 0.0 & 0.0 & 12.0 & 13.3\\
    Oryx-1.5-7B~\cite{liu2024oryx} & 20.5 & 13.5 & 14.8 & 10.1 & 3.5 & 15.1 & 13.8 \\
    VideoChat2~\cite{li2024mvbench}  & 36.2 & 13.7 & 12.5 & 2.5 & 1.0 & 17.0 & 20.3 \\
    Qwen2.5-VL-7B$^*$~\cite{bai2025qwen2} & 33.5 & 15.4 & 13.8 & 17.0 & 2.5 & 19.3 & 22.4\\
    InternVL-2.5-8B~\cite{chen2024internvl} & 44.2 & 8.7 & 7.8 & 0.7 & 0.1 & 17.6 & 24.9 \\
    Video-LLaMA3~\cite{zhang2025videollama}  & 41.9 & 23.0 & 23.1 & 0.9 & 0.2 & 21.7 & 27.0\\
    LLaVA-Video~\cite{zhang2024llava} & 49.5 & 10.5 & 12.2 & 1.9 & 1.3 & 20.8 & 27.3 \\
    VideoChat-R1.5~\cite{videochatr15} & 49.7 & 14.4 & 5.0 & 14.8 & 1.5 & 22.5 & 29.3 \\
    Open-o3-video$^*$~\cite{meng2025open} & 60.2 &  25.0 & 24.5 & 24.8  & \textbf{5.9} & 33.4 & 46.0 \\  \midrule
    \ours{} (Ours) & \textbf{61.1} &  \textbf{25.7} & \textbf{25.4} & \textbf{27.2}  & 5.3 & \textbf{34.3} & \textbf{47.5} \\
    $\Delta$ vs.\ Qwen2.5-VL-7B & +27.6 & +10.3 & +11.6 & +10.2 & +2.8 & +15.0 & +25.1 \\
    \bottomrule[0.15em]
  \end{tabular}
  }
  \label{tab:vstar}
\end{table*}

\subsection{Setup}

\noindent\textbf{Implementation Details.}
For \reasoner{} training, we follow a two-stage pipeline consisting of SFT-based cold-start initialization followed by RL, consistent with prior work~\cite{meng2025open,videor1,wang2025videorft}.
We first train the \selector{} and keep it frozen during the subsequent \reasoner{} training stage.
The \selector{} is trained using the training dataset of TVQA+~\cite{lei2020tvqa+} and VideoEspresso~\cite{han2025videoespresso}.
For \reasoner{} training, we adopt the same SFT and RL datasets as~\cite{meng2025open} to ensure a fair comparison.
For self-distillation, we select the top-2 samples among candidates with improved rewards and set the self-distillation weight to $\alpha=0.1$.
Additional implementation details, baseline details, and threshold ablations are provided in the Appendix.

\noindent\textbf{Benchmark and Metrics.}
We evaluate spatio-temporal reasoning on V-STAR~\cite{vstar}. 
For zero-shot general video understanding, we report results on WorldSense~\cite{hong2025worldsense}, VideoMMMU~\cite{videommmu}, PerceptionTest~\cite{perceptiontest}, and VideoMME~\cite{videomme}. 
For zero-shot temporal grounding, we use Charades-STA~\cite{gao2017tall}. 
For zero-shot spatio-temporal grounding, we use HCSTVG~\cite{tang2021human}. 
More details regarding metrics and benchmarks are in the Appendix.

\begin{table}[t]
\centering
\caption{\small \textbf{Performance across different general video understanding benchmarks.} * indicates our implementation performance. 
}
\resizebox{0.95\textwidth}{!}{
\scriptsize
\begin{tabular}{lccccccc}
\toprule[0.15em]
\textbf{Model} & 
\multicolumn{2}{c}{\textbf{VideoMME}} & 
\multicolumn{2}{c}{\textbf{WorldSense}} & 
\multicolumn{2}{c}{\textbf{VideoMMMU}} &
\textbf{PerceptionTest} \\
\cmidrule(lr){2-3} \cmidrule(lr){4-5} \cmidrule(lr){6-7} \cmidrule(lr){8-8} 
&\textbf{Overall} & Long  & \textbf{Overall} & Recognition & \textbf{Overall} & Perception & \textbf{Overall} \\
\midrule
\rowcolor{red!8} \multicolumn{8}{l}{\emph{Proprietary Models}} \\

GPT-4o & 71.9 & - & 42.6 & - & 61.2 & 66.0 & - \\
\midrule
\rowcolor{blue!8} \multicolumn{8}{l}{\emph{Tool-calling Methods}} \\
\rowcolor{gray!8} EgoR1 & - & \color{gray} 64.9 & - & - & - & - & - \\
\rowcolor{gray!8}  EgoR1 (w/o RAG) & \color{gray} 58.7 & \color{gray} 50.8 & \color{gray} 42.0 & \color{gray} 40.3 & \color{gray} 34.4 & \color{gray} 37.6 & \color{gray} 65.8 \\
\rowcolor{gray!8}  LongVT-RL-7B & \color{gray} 66.1 & - & \color{gray} 22.1 & \color{gray} 22.5 & \color{gray} 42.6 & \color{gray} 50.0 & \color{gray} 55.4 \\
\midrule
\rowcolor{green!8} \multicolumn{8}{l}{\emph{Tool-free Methods}} \\
Qwen2.5-VL-7B & 62.4 & 50.8  & 36.1 & 33.7 & 51.2 & 64.7 & 66.2 \\
VideoRFT-7B & 59.8 & 50.7 & 38.2 & 36.6 & 51.1 & 66.0 & - \\
VideoR1-7B & 61.4 &  50.6 & 35.5 & 32.8 & 52.4 & 65.3 & - \\
Open-o3-video-7B & 63.1* &  53.1* & 37.5 & 36.8 & 52.3 & 68.0 & 67.5 \\ \midrule
\ours{}-7B (Ours) & \textbf{63.3} & \textbf{53.2}  & \textbf{43.8} & \textbf{41.8} & \textbf{54.4} & \textbf{70.3} & \textbf{68.7} \\
\bottomrule[0.15em]
\end{tabular}
}
\label{tab:general_video_understanding}
\end{table}

\subsection{Main Results}
\textbf{Results on V-STAR.}
We first evaluate the spatio-temporal reasoning capability of \ours{} on V-STAR.
As shown in \cref{tab:vstar}, \ours{} consistently outperforms both proprietary and open-source baselines across all evaluation dimensions.
Compared to Qwen2.5-VL-7B, \ours{} improves VQA accuracy by \textbf{+27.6} (33.5→61.1) and overall performance by \textbf{+15.0} mAM and \textbf{+25.1} mLGM.
These results demonstrate that \ours{} significantly strengthens grounded spatio-temporal reasoning, yielding consistent improvements in answer correctness, temporal localization, and spatial grounding, thereby validating the effectiveness of grounding-aware RL with self-distillation.

\noindent\textbf{Results on general video understanding benchmarks.}
We compare the general video understanding capability of \ours{} with GPT-4o, tool-calling models, and tool-free baselines.
As shown in \cref{tab:general_video_understanding}, \ours{} consistently outperforms text-centric reasoning models (VideoR1 and VideoRFT) as well as tool-calling approaches, while remaining fully tool-free.
Notably, \ours{} shows substantial gains on perception-oriented categories, including WorldSense (Recognition) and VideoMMMU (Perception).

\begin{wraptable}{r}{0.45\linewidth}
\vspace{-10pt}
\centering
\scriptsize
\caption{\small \textbf{Zero-shot temporal grounding performance on Charades-STA.}}
\label{tab:temporal_grounding}
\resizebox{\linewidth}{!}{
\begin{tabular}{lcccc}
\toprule
 & R@0.3 & R@0.5 & R@0.7 & mIoU \\
\midrule
\rowcolor{yellow!8}\multicolumn{5}{l}{\textit{General Video LLMs}} \\
VideoChat & 9.0 & 3.3 & 1.3 & 6.5 \\
VideoLLaMA & 10.4 & 3.8 & 0.9 & 7.1 \\
Video-ChatGPT & 20.0 & 7.7 & 1.7 & 13.7 \\
Valley & 28.4 & 1.8 & 0.3 & 21.4 \\
\rowcolor{yellow!8}\multicolumn{5}{l}{\textit{Temporal Grounding Video LLM}} \\
GroundingGPT & - & 29.6 & 11.9 & - \\
Momentor & 42.6 & 26.6 & 11.6 & 28.5 \\
TimeChat & - & 32.2 & 13.4 & - \\
HawkEye & 50.6 & 31.4 & 14.5 & 33.7 \\
VTG-LLM & 51.2 & 33.8 & 15.7 & 34.4 \\
VTimeLLM & 51.0 & 27.5 & 11.4 & 31.2 \\
\rowcolor{yellow!8}\multicolumn{5}{l}{\textit{Grounded Reasoning Models}} \\
Open-o3-video & 62.6 & 45.6 & 24.5 & 42.5 \\
\ours{} (Ours) & \textbf{63.2} & \textbf{45.8} & \textbf{24.7} & \textbf{42.7} \\
\bottomrule
\end{tabular}
}
\vspace{-15pt}
\end{wraptable}
These results indicate that improved spatio-temporal grounding not only enhances localization quality but also translates into stronger video answering.

\noindent\textbf{Results on the temporal grounding benchmark.}
We evaluate \ours{} on a temporal grounding benchmark and compare it against existing VideoLLMs specialized in temporal grounding. 
As shown in \cref{tab:temporal_grounding}, \ours{} consistently outperforms all competing methods. 
These results indicate that the proposed visual prompt guidance effectively facilitates the learning of accurate temporal grounding during training.

\noindent\textbf{Results on the spatio-temporal grounding benchmark. } 
To further evaluate \ours{}, we conduct experiments on HCSTVG~\cite{tang2021human}, a benchmark requiring precise temporal localization and spatial grounding. 
As shown in \cref{tab:hcstvg_results}, \ours{} consistently outperforms existing tool-calling approaches and achieves substantial gains over Open-o3-video across all evaluation metrics. 
\begin{wraptable}{r}{0.5\textwidth}
  \centering
  \vspace{-0.3in}
  \caption{\small \textbf{Zero-shot spatio-temporal grounding performance on HCSTVG.}}
  \label{tab:hcstvg_results}
  \resizebox{0.48\textwidth}{!}{%
  \begin{tabular}{lccc}
    \toprule
    \textbf{Models} & \textbf{m\_vloU} & \textbf{vloU@0.3} & \textbf{vloU@0.5} \\
    \midrule
    \rowcolor{red!8} \multicolumn{4}{l}{\emph{Proprietary Models}} \\
    GPT-5.4 & 20.2 & 25.9 & 5.1 \\
    \midrule
    \rowcolor{blue!8} \multicolumn{4}{l}{\emph{Tool-calling Methods}} \\
    VideoChat-R1.5 & 11.3 & 11.9 & 2.3 \\
    VideoZoomer & 7.2 & 5.35 & 0.6 \\
    Conan & 6.9 & 5.0 & 0.7 \\
    Video-o3 & 3.8 & 3.6 & 0.3 \\
    \midrule
    \rowcolor{green!8} \multicolumn{4}{l}{\emph{Tool-free Methods}} \\
    Open-o3-video & 17.1 & 22.6 & 5.2 \\
    \ours{} (Ours) & \textbf{19.3} {\textcolor{blue}{(\textbf{+12.9\%})}} & \textbf{25.4} {\textcolor{blue}{(\textbf{+12.4\%})}} & \textbf{6.1} {\textcolor{blue}{(\textbf{+17.3\%})}} \\
    \bottomrule
  \end{tabular}%
  }
  \vspace{-0.15in}
\end{wraptable}

Notably, \ours{} remains competitive with GPT-5.4 despite being built on a significantly smaller open-source backbone.
These results suggest that the proposed visual coaching framework is beneficial for fine-grained spatio-temporal reasoning, where accurate localization of relevant objects and events is crucial. 

\subsection{Ablation Study}

\noindent\textbf{Ablation of \ours{}: input-adaptive visual prompting is crucial for grounded reasoning.}
As shown in~\cref{tab:ablation}, we progressively enable four components of \ours{}: (i) spatial grounding reward ($r_{\text{spa}}$), (ii) self-distillation loss ($\mathcal{L}_{\mathrm{SD}}$), (iii) fixed visual prompting (VP-F), and (iv) \selector (VP-S). While self-distillation slightly improves VQA accuracy, it does not consistently improve grounding performance. Fixed visual prompting further degrades grounding due to indiscriminate perturbations. In contrast, combining input-adaptive prompting with self-distillation achieves the best performance across all metrics, demonstrating the importance of selectively applying visual guidance. Furthermore, VP-guided training yields higher rewards and faster convergence (\cref{fig:analysis}c), suggesting that visual prompting provides effective grounding signals during RL training.
Additional ablation studies are provided in the Appendix.

\noindent\textbf{Statistics of adaptive visual prompting: adaptive visual prompting effectively targets difficult samples and provides strong learning signals.}
To better understand the behavior of \ours{}'s adaptive visual prompting mechanism, we analyze the distribution of designated sample difficulty (Easy vs.\ Hard), the visual prompts selected for hard samples by \selector, and the resulting reward gain from each prompt.
We compute the relative reward gain as $(R_i' - \bar{R}_i) / \bar{R}_i \times 100$.
As shown in \cref{fig:analysis} (a), 58\% of the samples are identified as hard.
For these hard samples, \selector dynamically chooses among multiple visual prompting strategies.
Importantly, applying these prompts consistently leads to reward improvements, with gains ranging from +56\% to +66\% across different prompt types.
These results indicate that visual prompting provides effective guidance for challenging instances and that the selector successfully identifies prompts that yield meaningful training signals.

\begin{figure}[t]
\centering
\begin{minipage}[t]{0.48\textwidth} 
\small \captionof{table}{\textbf{Ablation of \ours{} on V-STAR.} $r_{\text{spa}}$: object-aware spatial grounding reward. $\mathcal{L}_{\mathrm{SD}}$: self-distillation. VP-F: fixed visual prompting (darken). VP-S: \selector{}.}
\label{tab:ablation}
\footnotesize
\renewcommand{\arraystretch}{1.2}
\resizebox{\linewidth}{!}{
\begin{tabular}{c c c c | c c c}
\toprule[0.15em]
 \textbf{$r_{\text{spa}}$} & \textbf{$\mathcal{L}_{\mathrm{SD}}$} & \textbf{VP-F} & \textbf{VP-S} & \textbf{Acc} & \textbf{mAM} & \textbf{mLGM} \\
\midrule
 \ding{51} & -- & -- & -- & 59.4 & 31.1 & 42.7 \\
 \ding{51} & \ding{51} & -- & -- & 59.6 & 30.4 & 41.6 \\
\ding{51} & \ding{51} & darken & -- & 58.3 & 29.7 & 40.6 \\
\ding{51} & \ding{51} & - & \ding{51} & \textbf{60.7} & \textbf{31.3} & \textbf{43.1} \\
\bottomrule[0.13em]
\end{tabular}
}
\end{minipage}
\hfill
\begin{minipage}[t]{0.48\textwidth} 
\leavevmode\hbox{}
\captionof{table}{
\small \textbf{\selector as plug-and-play module.} 
We evaluate the zero-shot performance of combining \selector with different reasoners. 
}
\label{tab:vp_selector_alone}
\scriptsize
\resizebox{\linewidth}{!}{
\begin{tabular}{lcc}
\toprule[0.13em]

\textbf{Models}  
& \shortstack{\textbf{TVQA+} \\ \textbf{(In-domain)}} 
& \shortstack{\textbf{PerceptionTest} \\ \textbf{(Out-domain)}} \\

\midrule

Qwen2.5-VL-7B & 54.5 & 52.2 \\
+\selector & \textbf{56.2} & \textbf{56.7} \\ 
\midrule

GPT-4o & 71.8 & 61.5 \\
+\selector & \textbf{75.3} & \textbf{62.4} \\ 
\midrule

Gemini-2.5-Flash & 72.4 & 47.4 \\ 
+\selector & \textbf{76.3} & \textbf{50.4} \\

\bottomrule[0.15em]
\end{tabular}
}
\end{minipage}
\vspace{-10pt}
\end{figure}

\noindent\textbf{\selector{} as a plug-and-play module: adaptive visual prompting improves video reasoning across model-agnostic backbones.} 
As shown in~\cref{tab:vp_selector_alone}, we evaluate \selector{} as a standalone module by integrating it with various MLLM backbones to augment video inputs prior to reasoning.
We conduct experiments on the TVQA+ validation set (in-domain) and on a uniformly sampled subset from each category of the PerceptionTest validation set (out-of-domain).
Incorporating \selector{} consistently improves performance across different backbone models on both benchmarks.
These results demonstrate that \selector{} functions as a plug-and-play module for enhancing grounded visual reasoning.

\subsection{More Analysis}

\begin{figure}[t]
\centering

\begin{minipage}{0.38\linewidth}
\centering
\includegraphics[width=\linewidth]{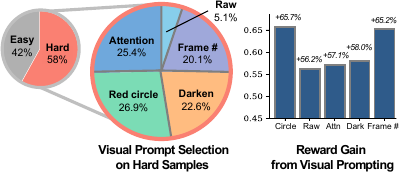}
\small (a) Statistics of input-adaptive visual prompting
\end{minipage}
\hfill
\begin{minipage}{0.28\linewidth}
\centering
\includegraphics[width=\linewidth]{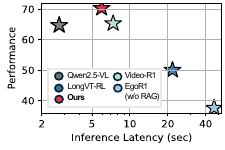}
\small (b) Comparison of inference latency
\end{minipage}
\hfill
\begin{minipage}{0.28\linewidth}
\centering
\includegraphics[width=\linewidth]{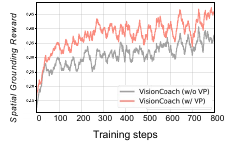}
\small (c) VP improves spatial grounding reward
\end{minipage}

\caption{\textbf{Additional analysis of \ours{}.}
We provide analysis including (a) statistics of visual prompting, (b) inference latency, and (c) the effect of visual prompting on spatial grounding reward.
}
\label{fig:analysis}
\end{figure}

\noindent\textbf{Inference latency: \ours{} achieves higher performance with lower inference latency.}
To evaluate per-sample inference latency, we measure runtime on a single NVIDIA RTX 6000 (40GB) GPU and compare text-centric reasoning, tool-calling reasoning, and \ours{}.
As shown in~\cref{fig:analysis} (b), \ours{} consistently outperforms both text-centric (Qwen2.5-VL, Video-R1) and tool-calling (EgoR1, LongVT-RL) baselines while operating at substantially lower inference latency than external tool-based approaches.

\begin{figure}[t]
    \centering

    \begin{minipage}[t]{\textwidth}
        \leavevmode\hbox{} 
        \centering
        \includegraphics[width=0.98\textwidth]{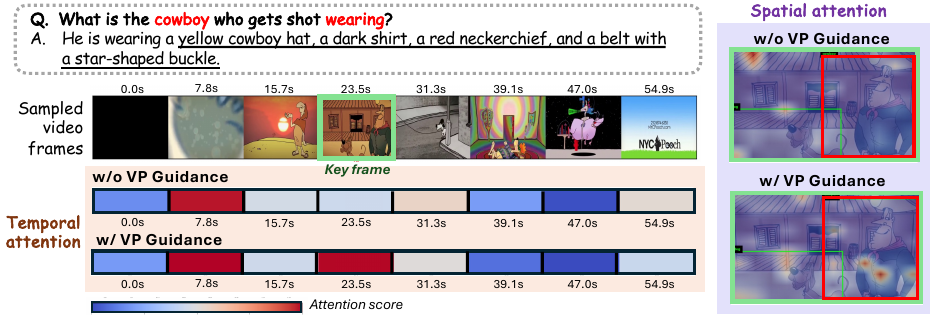}
        \caption{\small \textbf{Spatio-temporal attention map.} 
Visual prompting (VP) improves grounding in both temporal and spatial dimensions. The green box indicates the key frame, while the red box highlights the corresponding spatial region.
Temporally, VP increases attention on the correct key frame containing the cowboy.
Spatially, VP concentrates attention on the region corresponding to the queried visual attributes (e.g., the cowboy wearing specific clothing), while suppressing irrelevant regions. 
}
        \label{fig:attention}
    \end{minipage}
    \hfill

\end{figure}

\noindent\textbf{Spatio-temporal attention map: VP guidance encourages more focused temporal and spatial attention on visual evidence.}
To examine how visual prompting facilitates spatio-temporal grounding during training, we analyze the attention maps from the last layer of \ours{}. 
Specifically, we compute the average attention across all heads and measure the attention scores between the question tokens and visual tokens. 
Temporal attention is aggregated at the frame level, while spatial attention is aggregated over patch-level regions within each frame.
As illustrated in~\cref{fig:attention}, without VP guidance, both spatial and temporal attention are broadly dispersed across irrelevant frames and background regions, resulting in weak localization of the question-relevant event. 
In contrast, with VP guidance, the model concentrates attention on the key frame where the cowboy is shot and focuses on semantically relevant regions, such as the character and surrounding objects. 
This comparison indicates that VP guidance promotes more accurate temporal localization and spatial grounding, enabling better alignment between reasoning and visual evidence.
Additional analyses, including qualitative examples, reward ablations, and limitations are provided in the Appendix.

\section{Conclusion}
In this work, we introduced \ours{}, an instance-adaptive RL framework for grounded video reasoning. 
\ours{} applies visual prompting during RL to explicitly expose question-relevant evidence during learning using \selector and \reasoner. 
Experiments across multiple challenging benchmarks across tasks show consistent gains in both reasoning accuracy and grounding quality, while preserving a single, efficient inference pathway.

\section*{Acknowledgment}
This work was supported by ONR Grant N00014-23-1-2356, ARO Award W911\-NF2110220, DARPA ECOLE Program No. HR00112390060, and NSF-AI Engage Institute DRL2112635. The views contained in this article are those of the authors and not of the funding agency.

\newpage
\clearpage
\setcounter{page}{1}
\appendix
\section*{Contents of Supplementary Materials}

\noindent
\begin{tabular}{@{}ll}
\textbf{A. Evaluation Details} & \\
\hspace{1em} A.1. Baselines & \\
\hspace{1em} A.2. Evaluation Metrics & \\[0.5em]

\textbf{B. Implementation Details of \ours} & \\
\hspace{1em} B.1. Training Datasets & \\
\hspace{1em} B.2. More Training and Inference Details & \\[0.5em]

\textbf{C. Details of \selector} & \\
\hspace{1em} C.1. Details of Visual Prompting Types & \\
\hspace{1em} C.2. Data Collection & \\
\hspace{1em} C.3. Data Statistics & \\
\hspace{1em} C.4. Model Architecture & \\[0.5em]

\textbf{D. Additional Quantitative Results} & \\
\hspace{1em} D.1. Qwen3-VL Backbone & \\
\hspace{1em} D.2. \selector Training with Open-Source Data & \\
\hspace{1em} D.3. \selector Inference with Visual Prompting & \\
[0.5em]

\textbf{F. Additional Ablation} & \\
\hspace{1em} E.1. Ablation of Hyperparameter & \\
\hspace{1em} E.2. Ablation of Reward & \\
\hspace{1em} E.3. Ablation of Visual Prompting Assignment & \\
\hspace{1em} E.4. Ablation of Visual Prompting Combination & \\[0.5em]

\textbf{F. Visualization} & \\
\textbf{G. Limitation and Future Work} & 
\end{tabular}

\section{Evaluation Details}\label{sec:appendix:evaluation}

\subsection{Baselines}\label{sec:appendix:baseline}

We compare \ours{} with proprietary models (GPT-4o and Gemini-2.0-Flash) and open-source models including Video-LLaMA3~\cite{zhang2025videollama}, LLaVA-Video, and Open-o3-video~\cite{meng2025open}. 
We further include text-based reasoning models (VideoR1~\cite{videor1}, VideoRFT~\cite{wang2025videorft}), tool-calling models (Ego-R1-3B~\cite{egor1} with GPT-4o and Gemini-2.0-Flash, and LongVT-RL-7B~\cite{yang2025longvt}), and temporal grounding Video LLMs such as TimeChat~\cite{ren2024timechat}. 
For Ego-R1~\cite{egor1}, to enable evaluation across diverse benchmarks, we disable its dataset-specific RAG tools and retain the remaining tools.

\subsection{Evaluation Metrics}\label{sec:appendix:metric}

\textbf{V-STAR benchmark.} 
We evaluate \ours{}'s spatio-temporal grounding reasoning ability using the V-STAR~\cite{vstar} benchmark. 
V-STAR decomposes video reasoning into fine-grained Chain-of-Thought (CoT) questions along three dimensions: \textit{what}, \textit{when}, and \textit{where}. 
Chain1 corresponds to the reasoning order \textit{what–when–where}, while Chain2 follows \textit{what–where–when}. 
Each dimension is evaluated using dedicated metrics: answer accuracy (Acc), mean temporal IoU ($\text{m\_tIoU}$), and mean visual IoU ($\text{m\_vIoU}$), respectively.
To measure the overall reasoning performance across these dimensions, V-STAR first aggregates the three metrics using the Arithmetic Mean (AM) and Geometric Mean (GM):

\begin{equation}
AM = \frac{1}{3}(Acc + \text{m\_tIoU} + \text{m\_vIoU})
\end{equation}

\begin{equation}
GM = (Acc \times \text{m\_tIoU} \times \text{m\_vIoU})^{\frac{1}{3}} 
\end{equation}

However, when any metric becomes zero, GM collapses to zero. 
To mitigate this issue, the benchmark adopts a logarithmic transformation and defines the Logarithmic Geometric Mean (LGM):

\begin{equation}
LGM = -\frac{1}{3} \left[
\ln(1 - Acc + \epsilon) +
\ln(1 - \text{m\_tIoU} + \epsilon) +
\ln(1 - \text{m\_vIoU} + \epsilon)
\right]
\end{equation}
where $\epsilon$ is a small constant preventing $\ln(0)$ when any metric approaches 1.
Since the same questions can appear in different CoT chains, the benchmark further computes the mean AM (mAM) and mean LGM (mLGM) across chains:

\begin{equation}
mAM = \frac{1}{n}\sum_{k=1}^{n} AM_k,
\qquad
mLGM = \frac{1}{n}\sum_{k=1}^{n} LGM_k
\end{equation}
where $n$ denotes the number of reasoning chains. 
The mAM and mLGM metrics therefore capture the model's overall spatio-temporal reasoning performance across different reasoning orders.

\noindent\textbf{General video understanding benchmark.}
For all video understanding benchmarks, including VideoMME~\cite{videomme}, WorldSense~\cite{hong2025worldsense}, VideoMMMU~\cite{videommmu}, and PerceptionTest~\cite{perceptiontest}, we report multi-choice question accuracy. 

\noindent\textbf{Temporal grounding benchmark.} We evaluate temporal grounding performance using standard metrics on Charades-STA~\cite{gao2017tall}. 
Given a predicted temporal segment and the ground-truth interval for each query, we compute temporal Intersection over Union (tIoU) between the two segments. 
We report Recall at different IoU thresholds, including R@0.3, R@0.5, and R@0.7, which measure the percentage of queries whose predicted segment achieves a tIoU greater than the specified threshold. 
We also report mean IoU (mIoU), defined as the average temporal IoU across all queries.

\section{Implementation Details of \ours{} }\label{sec:appendix:details_implementation}

\subsection{Training Datasets}

In this section, we provide details for each SFT and RL stage. 

\noindent\textbf{Cold-start initialization.}
We use STGR-CoT-30k~\cite{meng2025open} dataset for SFT-based cold-start initialization.
This dataset is composed of four parts: 
(i) 4.1k temporal grounding chain-of-thought samples from TVG-ColdStart~\cite{tvg},
(ii) 5k spatial grounding samples from TreeVGR-SFT~\cite{treevgr},
(iii) 5.9k spatio-temporal samples curated from multiple video sources, including 3.9k samples from temporal grounding datasets such as ActivityNet~\cite{caba2015activitynet}, COIN~\cite{tang2019coin}, QueryD~\cite{oncescu2021queryd}, QVHighlight~\cite{qvhighlight}, and DiDeMo~\cite{didemo}, together with 2k samples from PLM-Rdcap~\cite{plmrdcap},
and (iv) 15k video reasoning samples from Video-R1-CoT~\cite{videor1}.

\noindent\textbf{Reinforcement learning.}
For RL, we train on the STGR-RL-36k  dataset~\cite{meng2025open}, which further increases the diversity of grounding scenarios.
The dataset contains four components: 
(i) 5.2k temporal grounding samples, including 2.3k from Time-R1~\cite{timer1} and 2.9k from TVG-RL~\cite{tvg},
(ii) 5k spatial grounding samples from VisCoT~\cite{viscot},
(iii) 10.9k spatio-temporal samples consisting of the 5.9k samples from Open-o3-video~\cite{meng2025open} and 5k filtered samples from VideoEspresso~\cite{han2025videoespresso},
and (iv) 15k video reasoning samples from Video-R1~\cite{videor1}.

\subsection{More Training and Inference Details}
We uniformly sample 16 frames from each video as input. 
For both cold-start and RL stage, we use one epoch with a learning rate of $1\times10^{-6}$. 
Unless otherwise specified, we use 8 NVIDIA GPUs with DeepSpeed ZeRO-3 optimization, a per-device batch size of 1, and gradient accumulation steps of 1. 
We set the maximum prompt length and completion length to 16{,}384 and 768 tokens, respectively. 
For GSPO training, we sample 4 rollouts per input and set the KL coefficient $\beta$ to 0.04. 
We use a cosine learning rate scheduler, weight decay of 0.01, and gradient clipping with a maximum norm of 5. For \ours{}, we empirically set the hard-question threshold to $k=2.21$, which means Open-o3-video's average number of rewards.
We use the average of $r_{spa}$, $r_{tmp}$, and  $r_{acc}$ to select self-distillation candidates $C_i$ and only use $r_{acc}$ to select the top 2 self-distilled examples.

\section{Details of \selector}\label{sec:appendix:details_selector}

\noindent\subsection{Details of Visual Prompting Types}\label{sec:appendix:selector_visualprompt} 

In this section, we provide details of each visual prompting type. 
We also include raw frames without visual prompting, as visual guidance is not always beneficial~\cite{api,zhang2025autov}.

\noindent\textbf{Key-object-based visual prompting (Darken/Red circle).}
We construct key-object-based visual prompts using object GT bounding boxes. 
For the red circle, we draw a circle centered at the bounding box center with a radius covering the object region to explicitly highlight the target object, following \cite{redcircle}. 
For the darken, we darken the regions outside the bounding boxes while preserving the object area, thereby guiding the model’s attention to the relevant object. 

\noindent\textbf{Frame numbering.}
To provide explicit temporal cues, we overlay frame indices on each video frame following \cite{numpro}. 
The key frame index is rendered as a red numeric label at the bottom-right corner. 

\noindent\textbf{Attention-based prompt.} 
Following API prompting~\cite{api}, we generate attention-based visual prompts using CLIP-ViT-L-14-336 by computing a query-conditioned spatial relevance map for each key frame. 
Visual features from an intermediate CLIP layer are projected onto the text embedding to obtain relevance maps, which are normalized and blended with the input frame to highlight query-relevant regions.

\begin{figure}[t]
    \centering
\includegraphics[width=0.99\columnwidth]{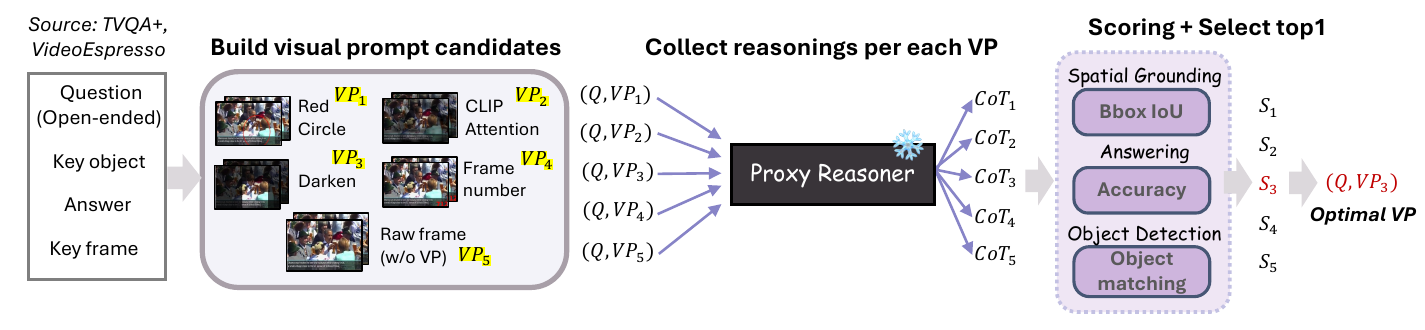}
\caption{\textbf{\selector training data generation pipeline. }
Given a video-question pair, we construct multiple visual prompting candidates, generate grounded reasoning with proxy reasoners, and derive pseudo-labels for selecting the most effective visual prompt.
}
    \label{fig:selector_data}
\end{figure}

\subsection{Data Collection}\label{sec:appendix:selector_data_collection}
We illustrate our data collection pipeline in \cref{fig:selector_data}. 
Given a video-question pair, we construct multiple visual prompting candidates, generate grounded reasoning with proxy reasoners, and derive pseudo-labels for selecting the most effective visual prompt.

\noindent\textbf{Proxy reasoner.}
For each visual prompting candidate constructed above based on TVQA+~\cite{lei2020tvqa+} and VideoEspresso~\cite{han2025videoespresso} training dataset, we generate grounded reasoning trajectories using multiple proxy reasoners, including Qwen3-VL-30B~\cite{bai2025qwen3}, Gemini-2.5-Flash~\cite{gemini2}, and GPT-4o~\cite{gpt4o}. 
Leveraging multiple proxy reasoners helps mitigate potential bias from any single model when estimating the effectiveness of different visual prompting strategies.
Specifically, we construct keyframes with different visual prompting strategies and provide each prompted keyframe sequence, together with the question, to the proxy reasoner. The proxy model is then prompted to produce a concise reasoning trace and a final answer, while explicitly grounding each reasoning step to spatio-temporal evidence in the frames. To enforce structured grounding, we require the reasoning to include object tags (\texttt{<obj>}), bounding box tags (\texttt{<box>}), and timestamp tags (\texttt{<t>}), which jointly specify the referenced object, its spatial location, and the corresponding moment in the video. Details of the proxy reasoner prompt are provided in \cref{fig:proxy_reasoner_prompt}.

\noindent\textbf{Scoring.}
Based on the proxy reasoner outputs, we aggregate their predictions to obtain an optimal visual prompt pseudo-label for each question. 
To select the most effective visual prompt for both grounded reasoning and answer prediction, we consider multiple signals, including answer correctness, spatial grounding quality, and object consistency. 
Specifically, we first parse the predicted bounding boxes, object names, and final answers from the generated reasoning trajectories. 
\begin{wrapfigure}{r}{0.35\linewidth}
\centering
\includegraphics[width=\linewidth]{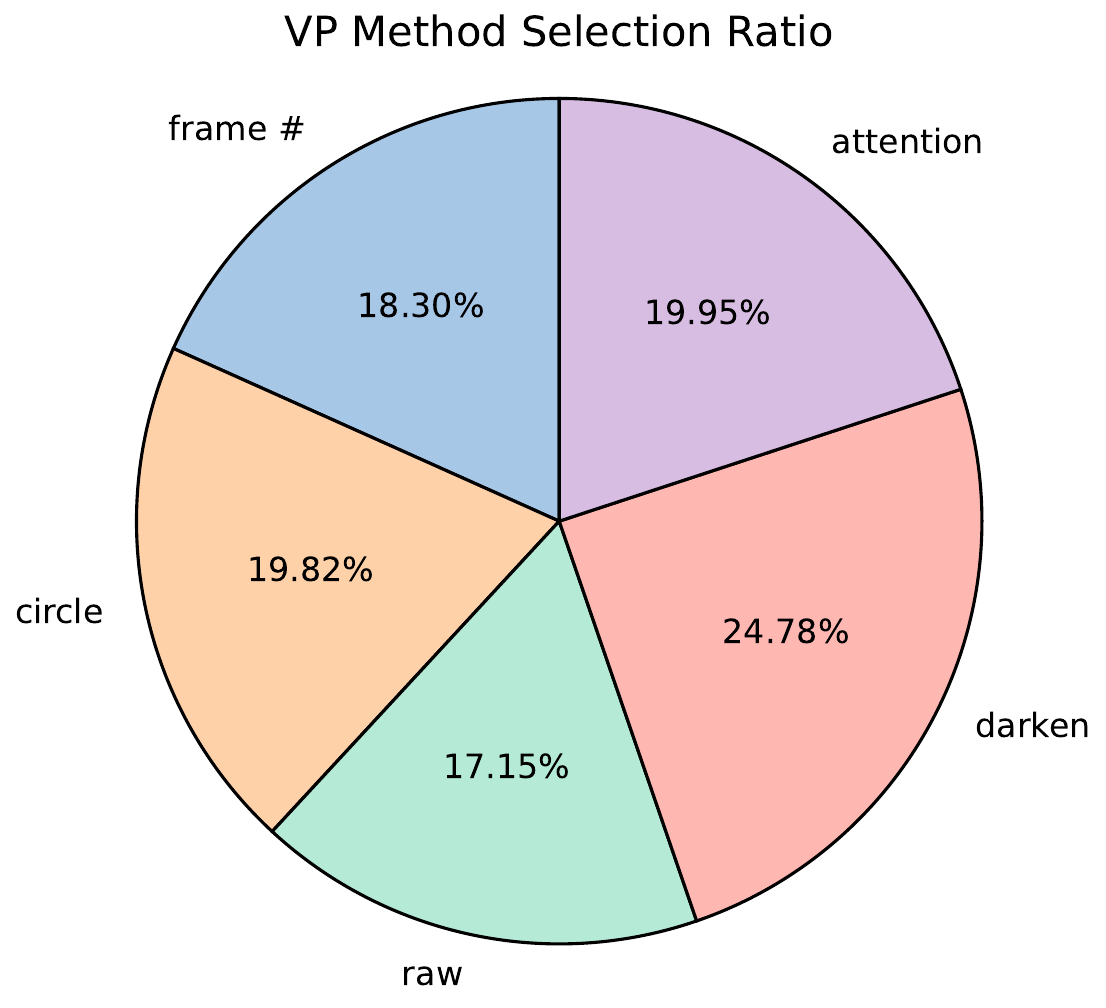}
\caption{\small \noindent\textbf{Distribution selected visual prompting of \selector training datasets.}}
\label{fig:vp_method_ratio}
\end{wrapfigure}
Answer quality is evaluated using accuracy for multiple-choice questions and ROUGE for open-ended answers. 
For spatial grounding, we compute the IoU between the predicted bounding boxes and the annotated boxes provided in the TVQA+ and VideoEspresso datasets. 
To measure object consistency, we compute an object matching score between predicted and ground-truth object names using a hierarchical similarity metric that considers exact match, substring match, fuzzy similarity, and word-overlap (Jaccard) matching.

\noindent\textbf{Pseudo-label selection.}
To aggregate results from the three proxy reasoners, we first filter visual prompting candidates where all three proxy reasoners predict the correct answer. 
Among these candidates, 
we select the optimal visual prompt by ranking them according to the average of the spatial IoU and object matching scores.

\subsection{Data Statistics}\label{sec:appendix:selector_data_stat}

We use 5,000 training samples from TVQA+~\cite{lei2020tvqa+} and 1,382 samples from VideoEspresso~\cite{han2025videoespresso}, both consisting of multiple-choice QA pairs with annotated keyframes and key-object bounding boxes. 
We additionally include 3,000 samples from the VideoEspresso training set containing open-ended question–answer pairs.
The distribution of selected visual prompting methods is illustrated in \cref{fig:vp_method_ratio}. 
This distribution indicates that explicit visual prompting is generally beneficial and that the optimal prompting strategy varies across samples.

\subsection{Model Architecture}\label{sec:appendix:selector_archi}
\noindent\textbf{Training.} 
We first formulate visual prompting selection as a generative prediction task without introducing an additional classification head.
Each sample consists of the question, optional key object, and video key frames, where multiple frames are combined into a single contact-sheet image.
We implement the \selector by fine-tuning Qwen2.5-VL-Instruct (3B) with LoRA~\cite{hu2022lora}, while freezing the backbone parameters.
We apply LoRA to the query, key, value, output, and MLP projection layers, with rank $r{=}16$, scaling $\alpha{=}32$, and dropout $0.05$.
We train the model with AdamW for $10$ epochs using learning rate $2\times10^{-4}$, weight decay $0.01$, and batch size $8$ per process.
The model is trained with token-level causal language modeling loss to generate one label from the discrete prompt set
$\mathcal{V}=\{\texttt{api\_prompt}, \texttt{circle}, \texttt{darken}, \texttt{numpro}, \texttt{raw}\}$.

\noindent\textbf{Inference.} 
At inference time, we use greedy decoding and map the generated text to the corresponding prompt class. 
To obtain the performance reported in \cref{tab:vp_selector_alone}, we first use \selector to predict the optimal visual prompt during inference, apply the selected visual prompting to the original input, and then feed it to each reasoner. 
If key objects are not provided as annotations (\eg, PerceptionTest), we extract them using Gemini-2.5-Flash~\cite{gemini2} and apply key-object-based visual prompting. Detailed prompts for key object extraction are provided in \cref{fig:gemini_keyobject}.

\section{Additional Quantitative Results}\label{sec:appendix:additional-quan-results}

\begin{table}[t]
\centering

\begin{minipage}[t]{0.48\linewidth}
    \centering
    \captionof{table}{\small\textbf{Inference with VP}}
    \label{tab:inference_vp}
    \resizebox{\linewidth}{!}{%
    \begin{tabular}{l|cc}
    \toprule
     & TVQA+ & PerceptionTest \\
    \midrule
    Qwen2.5-VL-7B & 54.5 & 52.2 \\
    \quad + Visual Prompting & 56.2 {\footnotesize\textcolor{blue}{(\textbf{+3.1\%})}} & 56.7 {\footnotesize\textcolor{blue}{(\textbf{+8.6\%})}} \\
    \midrule
    VisionCoach & 56.5 & 56.4 \\
    \quad + Visual Prompting & 56.7 {\footnotesize\textcolor{blue}{(\textbf{+0.4\%})}} & 57.1 {\footnotesize\textcolor{blue}{(\textbf{+1.2\%})}}\\
    \bottomrule
    \end{tabular}%
    }
\end{minipage}
\hfill
\begin{minipage}[t]{0.48\linewidth}
    \centering
    \renewcommand{\arraystretch}{1.1}
    \captionof{table}{\small\textbf{Qwen3-VL backbone}}
    \label{tab:qwen3_backbone}
    \resizebox{\linewidth}{!}{%
    \begin{tabular}{l|ccc}
    \toprule
     & VQA & mAM & mLGM \\
    \midrule
    Qwen3-VL-Instruct-8B & 32.0 & 18.1 & 21.1 \\
    Open-o3-video & 63.0 & 35.7 & 48.3 \\
    \rowcolor{yellow!8} VisionCoach (Ours) & 65.9 & 37.0 & 50.9 \\
    \bottomrule
    \end{tabular}%
    }
\end{minipage}

\vspace{.15in} 

\begin{minipage}[t]{0.48\linewidth}
    \centering
    \captionof{table}{\small\textbf{\selector training with open-source data
    }}
    \label{tab:opensource_vp}
    \resizebox{\linewidth}{!}{%
    \begin{tabular}{l|cc}
    \toprule
    \textit{Proxy Reasoner Data} & TVQA+ & PerceptionTest \\
    \midrule
    GPT-4o & 71.8 & 61.5 \\
    \rowcolor{yellow!8}\quad + Qwen3-VL & 74.2 & 61.9 \\
    \quad + GPT4o, Gemini-2.5 & 75.3 & 62.4 \\
    \midrule
    Gemini-2.5-Flash & 72.4 & 47.4 \\
    \rowcolor{yellow!8}\quad + Qwen3-VL  & 75.1 & 49.3 \\
    \quad + GPT4o, Gemini-2.5 & 76.3 & 50.4 \\
    \bottomrule
    \end{tabular}%
    }
\end{minipage}
\hfill
\begin{minipage}[t]{0.48\linewidth}
    \centering
    \renewcommand{\arraystretch}{0.81}
    \captionof{table}{\small\textbf{Ablation on visual prompt candidates}}
    \label{tab:vp_candidate_ablation}
    \resizebox{\linewidth}{!}{%
    \begin{tabular}{l|ccc}
    \toprule
    \textit{Visual Prompt Candidate} & VQA & mAM & mLGM \\
    \midrule
    \rowcolor{gray!8}Original \selector & 60.7 & 31.3 & 43.1 \\
    \midrule 
    frame\# & 57.3 & 29.1 & 39.8 \\
    circle & 57.7 & 29.3 & 39.9 \\
    darken & 58.3 & 29.7 & 40.6 \\
    darken + circle & 58.5 & 29.9 & 41.8\\
    darken + frame\# & 59.6 & 30.6 & 42.7 \\
    darken + frame\# + circle & 60.1 & 31.0 & 42.9\\
    \bottomrule
    \end{tabular}%
    }
\end{minipage}

\end{table}

\subsection{Qwen3-VL Backbone}
To evaluate the generality of \ours{} across different backbone models, we replace the \reasoner with Qwen3-VL-8B-Instruct and repeat the evaluation on V-STAR.
\cref{tab:qwen3_backbone} shows that \ours{} consistently outperforms Open-o3-video under the stronger Qwen3-VL backbone, improving VQA from 63.0 to 65.9, mAM from 35.7 to 37.0, and mLGM from 48.3 to 50.9.
These results demonstrate that the proposed visual coaching paradigm is not tied to a specific model family and transfers effectively to newer video-language backbones.

\subsection{\selector Training with Open-Source Data}
Although our default setting utilizes proprietary models to construct proxy reasoning data for \selector training, \ours{} is not fundamentally dependent on proprietary supervision.
We replace GPT-4o and Gemini-based proxy labels with Qwen3-VL-generated annotations and retrain \selector.
As shown in \cref{tab:opensource_vp}, the resulting selector still consistently improves performance across both TVQA+ and PerceptionTest.
Furthermore, combining open-source and proprietary supervision yields additional gains, suggesting that the proposed framework benefits from stronger proxy reasoners but remains effective under fully open-source settings.

\subsection{\selector Inference with Visual Prompting}
\ours{} is designed to leverage visual prompting during training while performing inference without additional prompts.
To verify whether visual prompting behavior is successfully internalized through self-distillation, we additionally evaluate \ours{} with visual prompting enabled at inference time.
As shown in \cref{tab:inference_vp}, the original Qwen2.5-VL model benefits substantially from visual prompting, achieving gains of +3.1\% on TVQA+ and +8.6\% on PerceptionTest.
In contrast, \ours{} obtains only marginal improvements when visual prompting is added during inference (+0.4\% and +1.2\%, respectively).
This result suggests that the proposed self-distillation effectively transfers visual prompting guidance into the model parameters, enabling prompt-free grounded reasoning at test time.

\section{Additional Ablation}\label{sec:appendix:additional-ablation}
In this section, we present various ablation on V-STAR~\cite{vstar} and VideoMME~\cite{videomme} using the spatio-temporal category subset of the STGR-RL dataset~\cite{meng2025open}. 

\subsection{Ablation of Hyperparameter}

\begin{table}[t]
\centering

\begin{minipage}{0.48\textwidth}
\centering
\caption{\small \textbf{Ablation of hard sample threshold ($k$).}}
\resizebox{\textwidth}{!}{
\scriptsize
\begin{tabular}{l | c c c}
\toprule[0.15em]
 \textbf{$k$}  & \textbf{Acc} & \textbf{mAM} & \textbf{mLGM} \\
\midrule
 10\%  & 58.7 & 30.4 &  41.6\\
 25\% &  57.0 & 29.7  & 40.0 \\
 50\% (default) & \textbf{60.7} & \textbf{31.3} & \textbf{43.1} \\
 75\% & 59.7 & 29.8 & 41.3 \\
\bottomrule[0.13em]
\end{tabular}
}
\label{tab:k_ablation}
\end{minipage}
\hfill
\begin{minipage}{0.48\textwidth}
\centering
\caption{\small \textbf{Ablation of self-distillation loss weight ($\alpha$).}}
\resizebox{\textwidth}{!}{
\scriptsize
\begin{tabular}{l | c c c}
\toprule[0.15em]
 \textbf{$\alpha$}  & \textbf{Acc} & \textbf{mAM} & \textbf{mLGM} \\
\midrule
 0 & 58.4 & 30.3 & 41.4 \\
 0.01 & 59.5 & 30.4 & 42.0 \\
 0.1 (default) & \textbf{60.7} & \textbf{31.3} & \textbf{43.1} \\
 0.5 & 60.5 & 30.9 & 42.6\\
\bottomrule[0.13em]
\end{tabular}
}
\label{tab:alpha_ablation}
\end{minipage}

\end{table}

In this section, we analyze key hyperparameters of our training framework, including the hard example threshold, the self-distillation loss weight, and the number of self-distillation candidates.

\noindent\textbf{Hard sample threshold.}
Since \ours{} selects hard samples based on a reward threshold, we analyze the effect of different threshold values in \cref{tab:k_ablation}. 
We determine the threshold based on the reward distribution of Open-o3-video~\cite{meng2025open} and set the mean reward as the default threshold. 
As shown in \cref{tab:k_ablation}, using a moderate threshold ($k=50\%$) achieves the best performance across all metrics. 
Lower thresholds include many relatively easy samples, providing weaker supervision, while overly high thresholds reduce the number of selected samples and limit effective training signals. 
These results suggest that selecting a balanced set of sufficiently challenging samples is important for effective training.

\noindent\textbf{Self-distillation loss weight.}
As shown in \cref{tab:alpha_ablation}, we analyze the effect of the self-distillation loss weight $\alpha$ in the overall objective. 
Introducing self-distillation already provides clear improvements over the baseline without distillation ($\alpha=0$), improving the answer accuracy from $58.4$ to $59.5$ even with a small weight ($\alpha=0.01$). 
The best performance is achieved at $\alpha=0.1$, which improves all metrics simultaneously, reaching $60.7$ Acc, $31.3$ mAM, and $43.1$ mLGM. 
However, further increasing the weight to $\alpha=0.5$ slightly degrades performance, suggesting that overly strong self-distillation may dominate the RL objective and reduce exploration during training. 
These results indicate that self-distillation is most effective when used as a moderate auxiliary signal that reinforces high-reward reasoning trajectories without overwhelming the primary GSPO optimization.

\noindent\textbf{Top-N self-distillation candidates.} 
As shown in \cref{tab:topn}, we analyze the effect of the number of trajectories used for self-distillation.
Selecting the top-2 candidates achieves the best performance across all metrics, while using only the top-1 candidate leads to slightly lower improvements due to reduced supervision signals.
However, increasing the number of candidates to top-3 degrades performance.
We hypothesize that including more candidates introduces \textit{lower-quality} reasoning trajectories into the distillation objective, which dilutes the supervision signal from the highest-reward trajectories.
As a result, the model may imitate suboptimal reasoning behaviors, weakening the benefits of self-distillation.
These results suggest that distilling a small set of high-quality trajectories provides a better balance between stable supervision and avoiding noisy imitation.

\subsection{Ablation of Reward}

In this section, we analyze the effect of different reward components and spatial reward designs in our training objective.

\noindent\textbf{Overall reward design ablation.}
As shown in \cref{tab:reward_ablation}, we ablate each reward component to analyze its individual effect.
Starting from the combination of format reward, temporal reward, and spatial reward (first row), adding answer reward improves all metrics, increasing accuracy from 56.9 to 58.5 while also improving grounding performance.
Further incorporating all reward components yields the best overall results, achieving 60.7 Acc, 31.3 mAM, and 43.1 mLGM.
These results show that each reward contributes complementary supervision, and that jointly optimizing answer correctness and grounded spatio-temporal reasoning is crucial for strong performance.

\noindent\textbf{Ablation of spatial reward design.} 
Prior spatial reward~\cite{meng2025open} typically encourages models to localize only a single region by assigning a reward based on the maximum IoU with any ground-truth box. 
However, grounded video reasoning often requires identifying multiple objects involved in the reasoning process.
To better support such behavior, we further ablate the effect of each component in our object-aware spatial grounding reward.
Using the same notation as our final reward, we first consider the spatial reward used in prior work~\cite{meng2025open}, which applies temporal gating but ignores object identity and assigns reward using only the maximum IoU with any ground-truth box:
\begin{equation}
r_{\text{spa}}^{\text{max}}(x,y)
=
\frac{1}{M}
\sum_{m=1}^{M}
\mathbf{1}\!\left[\Delta t_m \le \tau\right]
\max_{b^{\text{gt}}\in\mathcal{B}^{\text{gt}}(\tilde{t}_m)}
\mathrm{IoU}(b_m,b^{\text{gt}}).
\end{equation}

\noindent We then replace the max operator with an average over \textit{all} ground-truth boxes at the matched frame, yielding:
\begin{equation}
r_{\text{spa}}^{\text{avg}}(x,y)
=
\frac{1}{M}
\sum_{m=1}^{M}
\mathbf{1}\!\left[\Delta t_m \le \tau\right]
\frac{1}{|\mathcal{O}^{\text{gt}}(\tilde{t}_m)|}
\sum_{o\in\mathcal{O}^{\text{gt}}(\tilde{t}_m)}
\max_{b^{\text{gt}}\in\mathcal{B}^{\text{gt}}(\tilde{t}_m)}
\mathrm{IoU}(b_m,b^{\text{gt}}).
\end{equation}
\noindent Finally, our full object-aware spatial grounding reward further incorporates soft identity matching and averages IoU only over temporally aligned predictions with valid object matches:
\begin{equation}
r_{\text{spa}}(x,y)
=
\frac{1}{|\mathcal{I}|}
\sum_{m\in\mathcal{I}}
\frac{1}{|\mathcal{O}_m^{\text{match}}|}
\sum_{o\in\mathcal{O}_m^{\text{match}}}
\max_{b^{\text{gt}}\in\mathcal{B}^{\text{gt}}(\tilde{t}_m)}
\mathrm{IoU}(b_m,b^{\text{gt}}).
\end{equation}

\begin{table}[t]
\centering

\begin{minipage}{0.48\textwidth}
\centering
\caption{\small \textbf{Ablation on the number of top-N self-distillation candidates (rollout count: 4).}}
\resizebox{\textwidth}{!}{
\scriptsize
\begin{tabular}{l | c c c}
\toprule[0.15em]
 \textbf{Top $N$}  & \textbf{Acc} & \textbf{mAM} & \textbf{mLGM} \\
\midrule
 Top 1 & 59.0 & 30.1 & 41.3 \\
 Top 2 (default) & \textbf{60.7} & \textbf{31.3} & \textbf{43.1} \\
 Top 3 & 58.9 & 29.2 & 40.2 \\
\bottomrule[0.13em]
\end{tabular}
}
\label{tab:topn}
\end{minipage}
\hfill
\begin{minipage}{0.48\textwidth}
\centering
\renewcommand{\arraystretch}{1.4}
\caption{\small \textbf{Ablation of overall reward combination.}}
\resizebox{\textwidth}{!}{
\scriptsize
\begin{tabular}{c c c c | c c c}
\toprule[0.15em]
 \textbf{$r_{\text{fmt}}$} & \textbf{$r_{\text{acc}}$} & \textbf{$r_{\text{tmp}}$} & \textbf{$r_{\text{spa}}$} & \textbf{Acc} & \textbf{mAM} & \textbf{mLGM} \\
\midrule
\ding{51} & -- & \ding{51} & \ding{51} & 56.9 & 29.5 & 39.8 \\
\ding{51} & \ding{51} & \ding{51} & -- & 58.5 & 29.4 & 40.3 \\
\ding{51} & \ding{51} & \ding{51} & \ding{51} & \textbf{60.7} & \textbf{31.3} & \textbf{43.1} \\
\bottomrule[0.13em]
\end{tabular}
}
\label{tab:reward_ablation}
\end{minipage}
\end{table}

\begin{table}[t] 
\centering 
\renewcommand{\arraystretch}{1.1}
\caption{\small \textbf{Ablation of object-aware spatial reward components on V-STAR and VideoMME.} }
\resizebox{\textwidth}{!}
{ 
\scriptsize 
\begin{tabular}{l | c | c c | c c | c} 
\toprule[0.15em] 
 & \textbf{Acc} & \makecell{\textbf{Where} \\ \textbf{(Chain 1)}} & \makecell{\textbf{Where} \\ \textbf{(Chain 2)}} & \textbf{mAM} & \textbf{mLGM} & \textbf{VideoMME}\\ 
\midrule 
Open-o3-video ($r_{\text{spa}}^{\text{max}}$) & 58.4 & 25.4 & 4.6 & 30.3 & 41.4 & 61.9 \\ 
+ Average IoU ($r_{\text{spa}}^{\text{avg}}$) & 58.7 & \textbf{28.0} & 5.0 & 30.1 & 41.3 & 61.9 \\ 
+ Identity matching ($r_{\text{spa}}$) & \textbf{59.9} & 27.5 & \textbf{5.4} & \textbf{31.3} & \textbf{43.1} & \textbf{62.1} \\ 

\bottomrule[0.13em] 
\end{tabular} 
} 
\label{tab:spatial_reward_component} 
\end{table}

As shown in \cref{tab:spatial_reward_component}, replacing the max-IoU design with Average IoU improves grounding-related metrics, indicating that aggregating signals from multiple objects provides richer spatial supervision. 
Further incorporating identity matching leads to consistent gains across all benchmarks, confirming that enforcing object-level consistency is crucial for grounded reasoning. 
In contrast, directly averaging IoU across predicted boxes is less effective because many low-quality or irrelevant predictions dilute the supervision signal.

We further analyze how different spatial rewards affect grounding behavior in \cref{tab:num_obj_spatial_reward}.
The max-IoU reward from Open-o3-video~\cite{meng2025open} results in very few number predicted objects (0.25) and boxes (0.38) in predicted reasoning, reflecting a tendency toward single-region grounding.
Introducing Average IoU substantially increases the number of grounded objects (0.70) and boxes (1.18), indicating that multi-object supervision encourages richer spatial reasoning.
With identity matching, the model further improves both quantities while maintaining object-level consistency, demonstrating that our reward promotes more faithful multi-object grounding during reasoning.

\subsection{Ablation of Visual Prompting Assignment}

We evaluate the effectiveness of \selector during RL by replacing it with fixed visual prompting and a simple gating strategy. For the fixed setting, when a hard sample is detected, a predefined visual prompt is applied to the key frame. For the gating, different visual prompts are assigned based on reward signals: if the temporal reward is below the average, we apply frame numbering to the key frame; if the spatial reward is below the average, we apply darkening to the key region.
\begin{wraptable}{r}{0.5\linewidth}
\centering
\caption{\small \textbf{Average number of predicted objects and bounding boxes per generated reasoning.}}

\scriptsize
\renewcommand{\arraystretch}{1.3}
\begin{tabular}{l | c c}
\toprule[0.15em]
 & Avg.obj & Avg.box \\
\midrule
Open-o3-video ($r_{\text{spa}}^{\text{max}}$) & 0.25 & 0.38 \\
+ Average IoU ($r_{\text{spa}}^{\text{avg}}$) & 0.70 & 1.18 \\
+ Identity matching ($r_{\text{spa}}$) & \textbf{0.83} & \textbf{1.40} \\
\bottomrule[0.13em]
\end{tabular}

\label{tab:num_obj_spatial_reward}
\end{wraptable}
As shown in \cref{tab:vp_assignment}, applying a single prompt type to all hard samples (e.g., darken or circle) slightly degrades overall reasoning performance compared to the baseline, suggesting that different questions require different forms of visual guidance. 
The reward-based gating strategy improves grounding metrics (mAM and mLGM) and slightly boosts VideoMME performance, but its gains remain limited due to the coarse heuristic used for prompt assignment. 
In contrast, our learned \selector achieves the best performance across all metrics, improving both answer accuracy and grounding quality. These results highlight the importance of adaptive visual prompt selection conditioned on the input video and question for effectively guiding spatio-temporal reasoning during RL training.

\begin{table}[t] 
\centering 
\caption{\small \textbf{Ablation of visual prompting assignment to hard samples on V-STAR and VideoMME.} }
\resizebox{0.80\textwidth}{!}
{ 
\scriptsize 
\begin{tabular}{l | c c c | c} 
\toprule[0.15em] 
Gating method & \textbf{Acc} & \textbf{mAM} & \textbf{mLGM} & \textbf{VideoMME}\\ 
\midrule 
- & 59.6 & 30.4 & 41.6 & 61.9 \\ 
Fixed (darken) & 58.3 & 29.7 & 40.6 & 62.4 \\ 
Fixed (circle) & 57.7 & 29.3 & 39.9 & 62.0 \\ 
\midrule
Gating (darken, frame\#) & 59.6 & 30.6 & 42.7 & 62.6 \\ 
\midrule
\selector & \textbf{60.7} & \textbf{31.3} & \textbf{43.1} & \textbf{62.8} \\ 
\bottomrule[0.13em] 
\end{tabular} 
} 
\label{tab:vp_assignment} 
\end{table}

\subsection{Ablation of Visual Prompting Combination}
To understand the contribution of different visual prompting strategies, we ablate the visual prompt candidate pool used by \selector.
As shown in \cref{tab:vp_candidate_ablation}, using a single visual cue such as frame indices, circles, or darkening consistently improves grounding performance compared to the vanilla model.
Moreover, combining multiple visual prompts further improves performance, suggesting that different prompt types provide complementary grounding signals.
The full visual prompt combination achieves the best performance across all evaluation metrics, validating our design choice of constructing a diverse visual prompt candidate pool for adaptive selection.

\section{Visualization}\label{sec:appendix:reasoning_example}

\begin{figure}[t]
    \centering
\includegraphics[width=0.99\columnwidth]{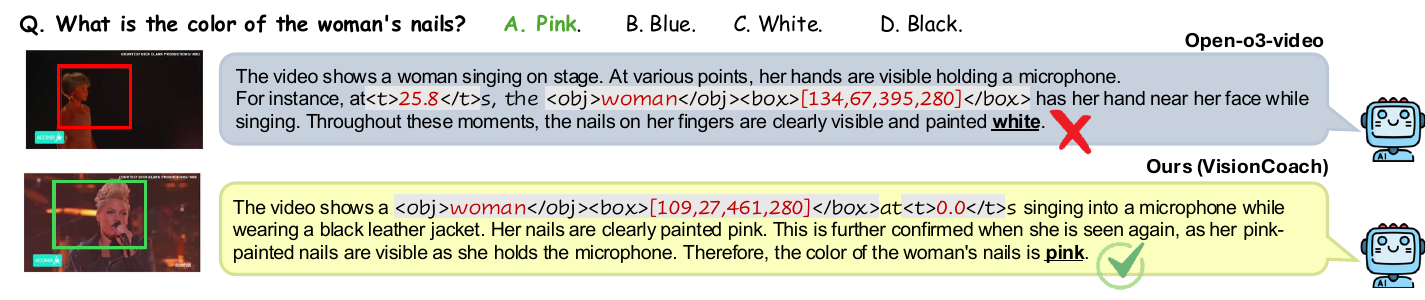}
\caption{\textbf{Comparison of grounding with Open-o3-video and \ours{}. }
}
    \label{fig:grounding_example}
\end{figure}

In~\cref{fig:grounding_example}, we compare grounded reasoning between Open-o3-video and \ours{}. Open-o3-video produces an incorrect answer (“white”) despite providing plausible explanations, indicating potential reliance on language priors rather than visual evidence. In contrast, \ours{} correctly identifies the nail color (“pink”) and supports the answer with precise spatial grounding on the woman’s hands across frames. 
This example illustrates that \ours{} encourages reasoning grounded with more reliable visual evidence, leading to more accurate predictions.

\section{Limitation and Future Work}\label{sec:appendix:limitation}
While \ours{} demonstrates strong improvements in grounded video reasoning, several limitations remain.
First, the framework still relies on training signals derived from grounding annotations (e.g., object boxes and temporal locations) to construct the spatial and temporal rewards. Although such annotations are available in many research benchmarks, scaling this approach to domains where precise grounding supervision is scarce may require additional strategies for obtaining reliable reward signals. Second, the effectiveness of \ours{} depends on the quality of the visual prompts selected during training. Although the VP-Selector learns to choose prompts adaptively, the prompt candidate pool is predefined and relatively simple (e.g., circles, darkening, frame indicators). More expressive forms of visual guidance or learned prompt generation may further improve the training signal for complex reasoning scenarios.
Future work may explore more efficient training strategies or scalable RL formulations to further improve practicality when applying the framework to larger video models and datasets.

\begin{figure}[h]
\centering
\begin{tcolorbox}[
    colback=gray!10,
    colframe=black!80,
    title={Prompt to the proxy reasoner
    },
    width=\columnwidth,
    boxrule=0.8pt
]
\scriptsize
\begin{verbatim}
You are a spatio-temporal video question answering model.

You will be given:
- A multiple-choice question with 5 options (A, B, C, D, E).
- K keyframes from a video in chronological order.
Some frames may contain visual prompts (e.g., boxes/circles/darkening/heatmaps 
or frame-number overlays).
Use these visual prompts as grounding hints when helpful.

Your job:
1) Produce a short, evidence-grounded reasoning that explicitly cites objects, 
bounding boxes, and timestamps using special tags.
2) Select the correct answer (A, B, C, D, or E).

OUTPUT FORMAT (MUST FOLLOW EXACTLY):
Return ONLY a valid JSON dictionary with exactly two keys: "reasoning" and "answer".
- "reasoning": a single paragraph string that includes grounding tags.
- "answer": a single string with the final answer (must be one of: "A", "B", "C", 
"D", or "E").

GROUNDING TAG REQUIREMENTS (MANDATORY):
- Object tag: <obj>...</obj>
- Box tag: <box>[x1, y1, x2, y2]</box>   
- Time tag: <t>...</t>                   (seconds)

Each grounding claim in the reasoning MUST include <obj>, <box>, and <t> together.
Use the exact substring pattern:
<obj>OBJECT</obj><box>[x1, y1, x2, y2]</box>at<t>TIME</t>s

Return format: 
{ "reasoning": "...<obj>...</obj><box>...</box>at<t>...</t>s...", "answer": "A" }

Rules:
- The reasoning must not exceed 200 words. 
- Use integer box coordinates in [0, 999].
- The answer must be exactly one letter: A, B, C, D, or E.
\end{verbatim}
\end{tcolorbox}
\caption{\textbf{Prompt to proxy reasoner.}}
\label{fig:proxy_reasoner_prompt}
\end{figure}

\begin{figure}[h]
\centering
\begin{tcolorbox}[
    colback=gray!10,
    colframe=black!80,
    title={Key object selection prompt by Gemini-2.5-Flash
    },
    width=\columnwidth,
    boxrule=0.8pt
]
\scriptsize
\begin{verbatim}
You are given a question, list of candidate visual objects and its answer about a video.

Your task is to identify the key visual object(s) that must be visually grounded
to make the answer correct. Select ONE primary visual object, and optionally up 
to TWO secondary visual objects.

[Question]
{question}

[Extracted objects]
{objects}

[Answer]
{answer}

Selection rules:
- Prefer selecting object names from [Extracted objects] if they reasonably 
match the question and answer. 
- Only select a new object NOT in [Extracted objects] if none of the extracted objects 
are reasonable.
- If there is NO specific object that needs to be grounded (e.g., abstract, scene-level,
or judgment-based questions),
  return "NONE" as the primary object.
- For "used_object", indicate whether each object (primary and secondary) comes from 
[Extracted objects]:
  - "yes" if the object is in [Extracted objects]
  - "no" if the object is newly selected (not in [Extracted objects]) or if primary 
  is "NONE"

Output JSON only in the following format:
{{
  "primary": "object name or NONE",
  "secondary": ["object name", ...],
  "used_object": {{
    "primary": "yes" or "no",
    "secondary": ["yes" or "no", ...]  // same length as secondary array
  }}
}}
\end{verbatim}
\end{tcolorbox}
\caption{\textbf{Prompt for extracting key object from question and answer pair.}}
\label{fig:gemini_keyobject}
\end{figure}

%
%
\clearpage
\bibliographystyle{splncs04}
\bibliography{main}

\newpage

\end{document}